\documentclass[lettersize,journal]{IEEEtran}
\usepackage{amsmath,amsfonts}
\usepackage{algorithmic}
\usepackage{algorithm}
\usepackage{array}
\usepackage[caption=false,font=normalsize,labelfont=sf,textfont=sf]{subfig}
\usepackage{textcomp}
\usepackage{stfloats}
\usepackage{url}
\usepackage{verbatim}
\usepackage{graphicx}
\usepackage{cite}
\usepackage{mathtools}
\usepackage{enumitem}

\DeclarePairedDelimiter\abs{\lvert}{\rvert}%
\DeclareMathOperator*{\argmax}{arg\,max}

\DeclareMathAlphabet\mathbfcal{OMS}{cmsy}{b}{n}
\renewcommand{\vec}[1]{\mathbf{#1}}

\usepackage[switch,columnwise]{lineno}
\begin{document}
	
	\title{gTLO: A Generalized and Non-linear Multi-Objective Deep Reinforcement Learning Approach}
	
	\author{Johannes Dornheim\\ %,~\IEEEmembership{Staff,~IEEE,}
		% <-this % stops a space
		\thanks{Johannes Dornheim is with the Institute for Applied Materials - Computational Materials Science (IAM - CMS), Karlsruhe Institute of Technology, Germany and the Intelligent Systems Research Group (ISRG), Karlsruhe University of Applied Sciences, Germany (e-mail: johannes.dornheim@mailbox.org).}% <-this % stops a space
		%\thanks{Manuscript received April 19, 2021; revised August 16, 2021.}
	}
	
	
	
	\maketitle
	
	\begin{abstract}
		%- Importance of Multi-Objective RL, 
		In real-world decision optimization, often multiple competing objectives must be taken into account. Following classical reinforcement learning, these objectives have to be combined into a single reward function. In contrast, multi-objective reinforcement learning (MORL) methods learn from vectors of per-objective rewards instead. 
		In the case of multi-policy MORL, sets of decision policies for various preferences regarding the conflicting objectives are optimized. This is especially important when target preferences are not known during training or when preferences change dynamically during application.
		% - advantage non-linear: optima in non-convex regions of the Pareto front can be reached by non-linear scalarization
		While it is, in general, straightforward to extend a single-objective reinforcement learning method for MORL based on linear scalarization, solutions that are reachable by these methods are limited to convex regions of the Pareto front. Non-linear MORL methods like Thresholded Lexicographic Ordering (TLO) are designed to overcome this limitation.
		% - advantage generalizing methods: can transfer knwoledge => sample efficiency, 	
		Generalized MORL methods utilize function approximation to generalize across objective preferences and thereby implicitly learn multiple policies in a data-efficient manner, even for complex decision problems with high-dimensional or continuous state spaces.
		% - contribution
		In this work, we propose \textit{generalized Thresholded Lexicographic Ordering} (gTLO), a novel method that aims to combine non-linear MORL with the advantages of generalized MORL. We introduce a deep reinforcement learning realization of the algorithm and present promising results on a standard benchmark for non-linear MORL and a real-world application from the domain of manufacturing process control.
	\end{abstract}
	
	%\linenumbers
	\section{Introduction}
	\IEEEPARstart{I}n real-world decision problems, decisions often result from balancing multiple objectives. In pure, single objective, reinforcement learning, the objectives are reflected by a single scalar reward function where the optimal decision is chosen to maximize the expected future reward \cite{sutton2018reinforcement}. If the objectives are not in alignment, the optimal decision is typically not clearly definable. Instead, the decision depends on scenario-specific preferences, e.g. the relative importance of each objective. Therefore, multi-objective reinforcement learning (MORL) is based on a setting in which the scalar reward function is replaced by a vector-valued reward function. This allows the design of reinforcement learning agents, which consider the various objectives during decision-making \cite{roijers2013}. In contrast to \textit{single-policy} MORL algorithms which use the reward vectors to optimize a specific policy for explicitly or implicitly defined objective preferences, \textit{multi-policy} MORL algorithms learn a set of preference-dependent optimal policies \cite{vamplew2011empirical}. This is especially useful in scenarios in which the online preferences are not known during agent training \cite{ruadulescu2020multi}. %Besides others \cite{hayes2021practical}, 
	This includes the \textit{dynamic preferences} scenario \cite{natarajan2005dynamic}, where the preferences are non-stationary during application. This scenario is given for example in domains in which the optimization criteria depend on dynamic prices of open markets \cite{ruadulescu2020multi}, or on changing requirements for process results \cite{dornheim2018multiobjective}. 
	
	In this work, we focus on \textit{multi-policy} MORL methods, which themselves can be classified into \textit{outer loop} methods and \textit{inner loop} methods \cite{roijers2017multi}. Outer loop methods decompose the multi-policy problem into a series of problems with defined preferences, that are solved by a single-policy approach. 
	Inner loop methods, on the other hand, are designed to solve the multi-policy optimization problem without the need for an outer loop. 
	A recent and broad review of MORL methods can be found in \cite{hayes2021practical}.
	
	One branch of inner loop MORL methods, which we name \textit{generalized} MORL, is to learn expected value functions that generalize over objective preferences \cite{castelletti2011multi, abels2019dynamic, friedman2018generalizing, yang2019generalized, tajmajer2018modular}. 
	Contrary to approaches where separate value functions are learned independently per preference, these \textit{generalized} MORL methods can transfer knowledge about the decision problem in between different preferences. %up to few-shot learning \cite{yang2019generalized}.
	The first published method that falls into this category is \textit{Multi-Objective Fitted Q-Iteration} (MOFQ) by Castelletti et al. \cite{castelletti2011multi}, a generalized MORL variant of \textit{Fitted Q-iteration} \cite{ernst2005tree} based on regression trees. 
	Friedman et al. \cite{friedman2018generalizing} took up this idea by extending \textit{Deep Deterministic Policy Gradients} \cite{lillicrap2015continuous} and \textit{Hindsight Experience Replay} \cite{Andrychowicz2017} to a generalized deep MORL method for continuous applications. 
	Abels et al. \cite{abels2019dynamic} proposed a generalized deep method based on generalized vector-valued \textit{Q-functions}, approximated by a weight-conditioned network (CN). They proposed \textit{Diverse Experience Replay} for sample efficient learning and to avoid catastrophic forgetting in the dynamic preferences setting. 
	Yang et al. \cite{yang2019generalized} proposed \textit{Envelope Q-learning} (EQL) based on a generalized form of the Bellman update equation that utilizes the convex envelope of the solution frontier for linear approaches. 
	Tajmajer \cite{tajmajer2018modular} proposed a method that can also be classified as generalized MORL, but differs from the already mentioned by introducing additional and dynamic per-state weights, which are called \textit{decision values}, and by learning an independent Deep Q-Network (DQN) \cite{mnih2015human} model per objective. \textit{Decision values} are learned by the agent as part of the extended DQNs and bind the preferences to the agent's state. 
	
	The ability of generalized MORL methods to efficiently optimize multiple preference-dependent policies has been shown in several publications, examples are \cite{castelletti2011multi, abels2019dynamic}. Castelletti et al. \cite{castelletti2011multi} showed that the sample efficiency of MOFQ compared to an \textit{outer loop} baseline rapidly increases with the number of policies for different preferences considered and already exceeds the baseline when more than five preferences are considered. 
	A key result of the comparative study in \cite{abels2019dynamic} is, that the proposed generalized MORL approach outperforms a state of the art outer loop approach based on \cite{mossalam2016multi} and \cite{natarajan2005dynamic} in terms of sample efficiency.
	
	% motivate non-linear MORL 
	Although current methods for generalized MORL are sample efficient, they combined preference values with expected values in a linear manner to derive decisions or to learn expected value functions. When observed in the space of per-objective expected returns, linear methods can identify policies that belong to expected returns from the \textit{convex coverage set}, which is a subset of the \textit{Pareto coverage set} \cite{vamplew2008limitations, roijers2013}. In practice, this limitation can lead to (i) agent behavior that is sub-optimal concerning the preference at hand, (ii) situations where balanced solutions are not found \cite{perny2010finding} and (iii) situations where small changes in the preference values cause huge changes in the policy \cite{vamplew2011empirical}. Non-linear MORL methods aim to overcome this problem. Basic approaches for non-linear MORL include 
	non-linear scalarization \cite{van2013scalarized}, 
	the explicit storing and pruning of sets of non-dominated Q-vectors \cite{van2014multi, ruiz2017temporal} and
	objective thresholding \cite{gabor1998multi, vamplew2011empirical, nguyen2020multi}.
	While early algorithms for non-linear MORL \cite{van2013scalarized, van2013hypervolume, ruiz2017temporal, gabor1998multi, vamplew2011empirical} are based on tabular reinforcement learning, more recent methods \cite{reymond2019pareto, nguyen2020multi, li2019urban} are based on deep reinforcement learning and are therefore also applicable in real-world applications with high-dimensional and often continuous state descriptions. The \textit{Pareto DQN} \cite{reymond2019pareto} is a deep MORL version of the \textit{Pareto Q-Learning} (PQL) algorithm \cite{van2014multi}. It aims to directly approximate the Pareto front and, to the best of our knowledge, is the only published approach that aims to solve multi-policy MORL problems in a non-linear and \textit{inner loop} fashion.
	
	% TLO
	\textit{Thresholded Lexicographic Ordering} (TLO) is a popular non-linear single-policy approach, proposed by G\'{a}bor et al. \cite{gabor1998multi} for multi-objective problems, in which one of the reward components has to be optimized, while the other components are constrained by minimum thresholds. 
	Vamplew et al. \cite{vamplew2011empirical} proposed an alternative implementation of TLO based on vector-valued multi-objective Q-Learning, \textit{Thresholded Lexicographic Q-learning} (TLQ). They showed that TLO / TLQ can be used as the basis of outer loop non-linear multi-policy learning by varying the minimum thresholds between the single-policy optimization runs. TLO / TLQ is applicable as long as the problem is finite-horizon and the reward components, that are subject to constraints are non-zero on terminal states only \cite{vamplew2011empirical}.
	Nguyen et al. \cite{nguyen2020multi} combined TLQ with DQN for deep reinforcement learning tasks with graphical state representations. Li et al. \cite{li2019urban} proposed a single-goal algorithm with adaptive thresholding and Q-function factoring methods for autonomously navigating intersections based on DQN and TLO.
	
	% contribution
	\begin{figure}
		\centering
		\includegraphics[width=0.7\linewidth]{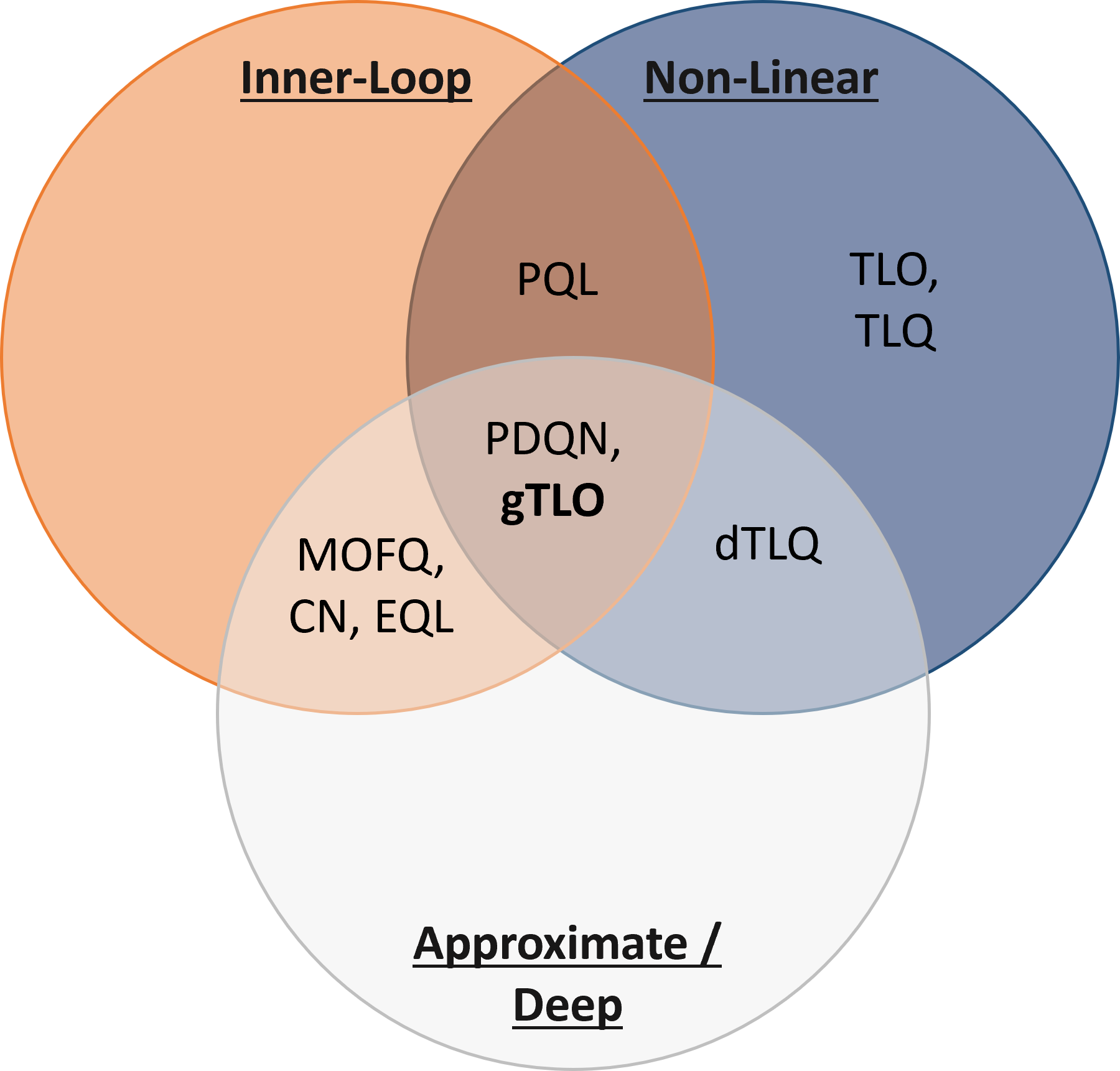}
		\caption{Grouping of the most related multi-policy MORL-methods discussed above by various aspects represented as a Venn Diagram. Shown are: Thresholded Ordering methods (TLO \cite{gabor1998multi}, TLQ \cite{vamplew2011empirical}, \textit{dTLQ} \cite{nguyen2020multi}), generalized MORL methods (MOFQ \cite{castelletti2011multi}, CN \cite{abels2019dynamic}, EQL \cite{yang2019generalized}), methods that directly approximate the Pareto-front (PQL \cite{van2014multi}, PDQN \cite{reymond2019pareto}) and \textit{gTLO} (ours). }
		\label{fig:morl_taxonomy}
	\end{figure}
	
	In this paper we present a MORL method that is (a) non-linear, (b) solves multi-policy problems in a generalized and thus inner loop fashion, and (c) can be combined with value-based deep reinforcement learning. The most related of the methods identified and discussed above are grouped by these three aspects in the Venn Diagram in Fig. \ref{fig:morl_taxonomy}.
	As discussed above, non-linear methods have the advantage that they are not limited to finding solutions from the \textit{convex coverage set} in general. Current generalized MORL approaches (including MOFQ \cite{castelletti2011multi}, CN \cite{abels2019dynamic}, EQL \cite{yang2019generalized}) fall into the linear regime.
	While tabular methods (e.g. PQL \cite{van2014multi}, TLO \cite{gabor1998multi} and TLQ \cite{vamplew2011empirical}) are able to solve small-scale problems, approximate reinforcement learning and deep reinforcement learning enables to scale to real-world problems with highly-dimensional or continuous state spaces without suffering the \textit{curse of dimensionality}.
	Most current non-linear multi-policy methods, including the TLO-based methods (TLO \cite{gabor1998multi}, TLQ \cite{vamplew2011empirical}, \textit{dTLQ} \cite{nguyen2020multi}), rely on a simple outer loop strategy and determine multiple policies by independently running a single-policy algorithm per preference. While this approach is limited to cases with only a few optimal policies and with knowledge about the location and course of the Pareto front (\cite{vamplew2011empirical, nguyen2020multi}), inner loop methods enable to scale to more realistic cases with many Pareto-optimal solutions and without knowledge about the location of those.
	The only published method we found that falls into the inner intersection of the Venn diagram is PDQN \cite{reymond2019pareto}, which is an "initial attempt" to integrate PQL into deep RL \cite{hayes2021practical}. However, presented results of PDQN indicate that the method in its current form is limited to problems with low-dimensional state spaces and only delivers very raw estimates of the true \textit{Pareto coverage set} \cite{reymond2019pareto}.
	
	The contribution of our work is threefold:
	\begin{enumerate}
		\item We propose gTLO: a non-linear generalized MORL method that combines the ability of generalized MORL to learn multiple optimal policies in a sample efficient manner, with the ability of the non-linear MORL method TLO to not being limited to the \textit{convex coverage set}. To the best of our knowledge \textit{gTLO} is the first non-linear generalized MORL method.% and an alternative and the first approximate non-linear inner loop method besides Pareto DQN \cite{reymond2019pareto}.	
		\item While \textit{gTLO} is combinable with various value-based reinforcement learning algorithms, we present and evaluate an implementation based on vanilla \textit{Deep Q-networks} (DQN).
		\item We present a comprehensive evaluation of our algorithm on two environments. First, we use an image-based version of the deep-sea treasure \cite{vamplew2008limitations} environment, which is a common benchmark, especially for non-linear MORL. Second, we evaluate our method on an environment from the manufacturing context: the multi-objective online optimization of blank holding forces of deep drawing processes \cite{dornheim2020model}.
	\end{enumerate}
	
	\section{Background}\label{background}
	
	\subsection{Single Objective Reinforcement Learning}
	\label{background_rl}
	
	The formal basis of reinforcement learning (RL) is the Markov decision process (MDP) \cite{sutton2018reinforcement}. MDPs are described by the tuple $(S,A,R,\gamma,P)$, where actions $a\in A$ lead from a Markovian state $s\in S$ to a subsequent state $s'\in S$, following the state transition probability function $P(s'\vert s, a)$. A reward value $r\in\mathbb{R}$ is given to the agent during the transition from $s$ to $s'$, following the reward function $R(s,a,s')$. In some processes, especially in the case of infinite-horizon MDPs, future rewards are discounted by a discount factor $\gamma\in [0,1]$. 
	We limit ourselves to finite-horizon MDPs within this work, where special absorbing states $\bar{S}\subset S$ exist. One sequence of decisions leading from a starting state $s_0$ to one of the absorbing states $\bar{S}$ is called an episode. Reinforcement learning methods aim to learn a policy $\pi:S\times A\rightarrow [0,1]$ during interaction with the process without any given a priori knowledge of $P$ or $R$. When following $\pi$, for any state $s$ the action $a$ is selected according to the probability $\pi(s,a)$. The goal is to learn an optimal policy $\pi^*$, which maximizes the expected future rewards collected until an absorbing state is reached by the agent. As usual in value-based RL, we consider deterministic optimal policies $\bar{\pi}^*$ \footnote{When $R$ is bounded, $A$ is finite and if at least one optimal policy exists for the MDP, it is guaranteed that at least one deterministic optimal policy $\bar{\pi}^*$ exists for that MDP (compare \cite{puterman2014markov}, Ch. 6). For multi-objective MDPs (MOMDP), scenarios exist where stochastic policies can dominate deterministic policies \cite{roijers2013}. In the case of a finite horizon, this can be prevented by extending the MOMDPs state space.}. We denote deterministic policies by $\bar{\pi}$ throughout the paper and define it as a direct mapping from states to actions $\bar{\pi}:S\rightarrow A$.
	
	In value-based reinforcement learning, functions of expected, possibly discounted, future rewards are learned. The $Q$-function is such a function and is defined for any policy $\pi$ on state-action tuples by
	
	\begin{equation}
		\label{q_function}
		Q_\pi(s,a)= \mathbb{E}_{P, \pi}\big[R(s,a,s') + \gamma Q_\pi(s',a')\big].
	\end{equation}
	
	The Q-function of an optimal policy $\bar{\pi}^*$ is denoted by $Q^*=Q_{\bar{\pi}^*}$. In Q-learning \cite{watkins1989learning}, $Q^*$ is learned by iteratively applying the Bellman Q-function update
	
	\begin{equation}
		\label{q_learning_update}
		Q(s,a)\leftarrow Q(s,a) + \alpha\tau
	\end{equation}
	
	based on experience tuples $(s,a,s',r)$, where 
	
	\begin{equation}
		\tau:=r + \gamma \max_{a' \in A} Q(s',a')-Q(s,a)
	\end{equation}
	is the so called \textit{temporal difference error}. For online learning, experience tuples are collected during process interaction by following an explorative policy $\pi_\epsilon$. As long as for all actions $a\in A$ and states $s\in S$ it is true that $\pi_\epsilon(s,a)>0$, the estimation $Q$ is guaranteed to converge to $Q^*$ \cite{sutton2018reinforcement}. Once $Q^*$ is given, an optimal deterministic policy can be easily extracted: $\bar{\pi}^*(s)=\argmax_{a\in A}Q^*(s,a)$.
	
	Pure Q-learning relies on an $\mathbb{R}^{\vert S\vert\times \vert A\vert}$ matrix to store the Q-value estimations, thus it suffers the \textit{curse of dimensionality} regarding the state space dimensions and can only be applied to discrete state spaces. In real-world applications, where a state $s$ is often described by a real-valued vector $\vec{s}\in\mathbb{R}^n$ or matrix, function approximation is used to overcome the problems of tabular learning. In value-based deep reinforcement learning, approaches are typically based on the \textit{Deep Q-Networks} algorithm (DQN) \cite{mnih2015human}. In DQN, the $Q$-function is approximated by a neural network $\mathcal{Q}\approx Q$. The approximation is usually implemented as $\mathcal{Q}(\vec{s}, \theta)$, where $\vec{s}$ is a description of the state $s$, $\theta$ are the network parameters and $\mathcal{Q}(\vec{s},\theta)_i$ approximates $Q(s,a_i)$, where actions are indexed by $i$. For the sake of simplicity, we instead use the notation $\mathcal{Q}(\vec{s},a,\theta)\approx Q(s,a)$ throughout the paper. %In the common deep reinforcement learning case where the state is described by an image in the form of $c$ channels of $x\times y$ features ($\vec{s}\in \mathbb{R}^{(x,y,c)}$), convolutional layers are used as part of $\mathcal{Q}$.
	
	Experience-tuples are stored in a so-called \textit{replay memory}, from which mini-batches of experiences are uniformly sampled for parameter updates. The training target for an experience tuple $(s,a,s',r)$ is then

	\begin{equation}
		\label{dqn_target}
		y_\mathcal{Q}\leftarrow r + \gamma \max_{a'\in \vec{A}} \mathcal{Q}(\vec{s'},a',\theta^-),
	\end{equation}
	
	where the \textit{target-parameters} $\theta^-$ are decoupled from the continuously updated \textit{online-parameters} $\theta$, that are used to form the explorative policy $\pi_\epsilon$ (and thereby influence the distribution of sampled experience tuples) to stabilize learning. Every $n_\theta$ time steps the target-parameters are updated by replacement $\theta^-\leftarrow\theta$.

	\subsection{Multi-Objective Reinforcement Learning}
	\label{background_morl}
	
	%\textbf{General / Linear}
	In many decision optimization problems, instead of a single objective, multiple objectives have to be optimized simultaneously. These problems can be formalized as multi-objective Markov decision processes \cite{roijers2013} (MOMDPs) ($S,A,P,\mathcal{R},\gamma$). Instead of scalar rewards, $\mathcal{R}(s,a,s')$ emits a reward vector $\vec{r}\in\mathbb{R}^I$, where each component reflects one of the $I$ objectives. The system dynamics are described by Markov states $s\in S$, actions $a\in A$, and the transition probability function $P$ as in the standard MDP. %In some MOMDP definitions also the discount factor depends on the objective, for our purposes we stick to the scalar discount factor $\gamma$. 
	In typical MOMDPs, the objectives are not in alignment, and instead of a single policy that solves the process, the optimal policy depends on the importance of each objective, according to the users, or the application-dependent preference. Multi-objective reinforcement learning (MORL) methods differ in how these preferences are expressed. 
	
	One can quantify the expected multi-objective return of each policy when starting in state $s$ on time step $t$ by defining a vector-valued multi-objective state-value function
	
	\begin{equation}
		\label{state_value}
		\vec{V}_\pi(s) = \mathbb{E}_{P, \pi}\big[\sum_{k=t}^\infty\gamma^k \vec{r}_{k+1} \big],
	\end{equation}
	
	where for the considered case of finite-horizon MDPs all rewards are zero ($\vec{r}=\vec{0}$) after the end of the episode. When considering a finite-horizon MOMDP with a fixed starting-state $s_0$, the expected multi-objective return per episode is given by $\vec{V}_\pi(s_0)$, which we denote by $\vec{V}_\pi$. Policies $\pi$ that are optimal for any trade-off between the objectives are part of the so-called \textit{Pareto front}. A policy $\pi\in\Pi$ is part of the Pareto front if it is not dominated by any other policy $\pi'\in\Pi\setminus \pi$ through the \textit{Pareto dominance relation} $(\forall i:\vec{V}_{\pi',i}\geq \vec{V}_{\pi,i})\land(\exists i:\vec{V}_{\pi',i} > \vec{V}_{\pi,i})$ for $1\leq i \leq I$, where $\Pi$ is the set of executable policies. While various policies $\pi$ on the Pareto front may have the same expected return value $\vec{V}_\pi$, one is typically only interested in a single policy per value. A so-called \textit{Pareto coverage set} (PCS) as desired solution set, therefore, is a subset of the Pareto front and consists of one policy per undominated value \cite{roijers2013}. 

	In MORL, preferences for $I$ objectives are often expressed by parameters $\vec{w}\in\mathbb{R}^I$ of a scalarization function $f(\vec{r},\vec{w})$ and the goal is to maximize the \textit{scalarized expected return} $f\big(\vec{V}_\pi(s),\vec{w}\big)$. Due to its simplicity, usual choice for scalarization is the weighted sum
	
	\begin{equation}
		\label{weighted_sum_scalarization}
		f_\text{s}(\vec{r},\vec{w})=\sum_{i\in[1,m]} \vec{r}_i\vec{w}_i,
	\end{equation}
	
	which is additive 
	($f_\text{s}(\vec{r},\vec{w})+f_\text{s}(\vec{r}',\vec{w})=f_\text{s}(\vec{r}+\vec{r}',\vec{w})$) 
	and the \textit{scalarized expected return} (SER) equals the \textit{expected scalarized return} (ESR) 
	
	\begin{equation}
		\label{linear_ser_esr}
		f_\text{s}\Big(\mathbb{E}_{P, \pi}\big[\sum_{k=t}^\infty\gamma^k \vec{r}_{k+1} \big],\vec{w}\Big) = \mathbb{E}_{P, \pi}\big[\sum_{k=t}^\infty\gamma^k f_\text{s}(\vec{r}_{k+1}, \vec{w}) \big].
	\end{equation}
	
	According to this property, linear scalarization is easily combine-able with conventional value-based reinforcement learning methods. In the most basic case, Q-learning can be used to learn the action-dependent ESR \cite{castelletti2002reinforcement}.

	% ====================================== generalized MORL ===========================================
	Also, existing \textit{generalized MORL} methods are based on linear scalarization. Generalization across preference-vectors $\vec{w}$ is realized 
	either 
	\begin{itemize}
		\item by learning a generalized function of the \textit{expected scalarized return} (ESR) $\mathcal{Q}(\vec{s},a,\vec{w},\theta)\approx Q_\vec{w}(s,a_i)$ \cite{castelletti2011multi, friedman2018generalizing}, where $Q_\vec{w}$ is the Q function for $f_\text{s}(\bullet,\vec{w})$, or
	 	\item by learning a vector-valued Q-function $\mathbfcal{Q}(\vec{s},a,\vec{w},\theta)\approx\vec{Q}(s,a,\vec{w})$, where $\vec{Q}(s,a,\vec{w})_i$ is the expected future reward for objective $i$ and preference $\vec{w}$, and scalarizing it (SER) for policy-extraction \cite{yang2019generalized,abels2019dynamic}. 
	\end{itemize}
	
	% ====================================== non-linear (generalized) ===========================================
	Policies that are optimal w.r.t. $f_\text{s}$ for any preference-vector $\vec{w}$ form the \textit{convex coverage set} (CCS), which is a subset of the \textit{Pareto coverage set} \cite{vamplew2008limitations, roijers2013}. The CCS includes only policies $\pi$, where the expected return $\vec{V}_{\pi}$ is located in a convex region of the Pareto front. Consequently, linear MORL is limited to identifying PCS policies with expected returns in non-convex regions, and thereby incomplete \cite{perny2010finding,vamplew2011empirical}. This motivates the research of non-linear approaches.
	 
	% non-linear morl
	Unlike linear scalarization, non-linear scalarization functions are not additive and the \textit{scalarized expected return} (SER) in general, differs from the \textit{expected scalarized return} (ESR) \cite{perny2010finding, hayes2021practical}. The additivity of reward sequences is a basic assumption of the Bellman update. As this is not the case for sequences of non-linearly scalarized rewards, policies that are optimal under ESR learning, (i.e. following $\max_a \mathcal{Q}(\vec{s},a,\vec{w},\theta)$), in general, differ from policies that are optimal under the SER criterion (i.e. following $\max_a f\big(\vec{\mathbfcal{Q}}(\vec{s},a,\vec{w},\theta)\big)$). The objective in multi-objective optimization is typically to maximize the \textit{scalarized expected return} (SER) $f(E(s_0))$ and to find the \textit{Pareto coverage set}\footnote{There are cases in which the maximization of \textit{expected scalarized returns} is more suitable \cite{roijers2013}. This case is covered in recent work \cite{roijers2018multi}.}.

	\subsection{Thresholded Lexicographic Ordering}
	\label{TLO}
	
	%TLO (TLQ) in detail
	Instead of a \textit{scalarization function}, Thresholded Lexicographic Ordering (TLO) \cite{gabor1998multi} is based on a non-linear action selection mechanism. Preferences are given in the form of minimum thresholds $\vec{t}=(\vec{t}_0,...,\vec{t}_I)^\top$ for the $I$ reward components ($\vec{r}\in\mathbb{R}^I$). The last objective is unconstrained, thus formally $\vec{t}_I=+\infty$. TLO aims to select an action that maximizes $\vec{Q}_I$, while the minimum constraints $\vec{Q}_i>\vec{t}_i$ are met for $i\in[0,I-1]$. During learning, the action selection mechanism guides the agent to gradually find policies that fulfill the threshold constraints. The action $a^*$ is selected by TLO in state $s$, if the superior-proposition $\text{sup}(a^*, a, s, 0)$ is \textit{true} for every $a\in A\setminus a^*$ \cite{vamplew2011empirical}. Based on a thresholded Q-function
	
	\begin{equation}\label{qt}
	\vec{Qt}(s,a)_i=\text{min}(\vec{Q}(s,a)_i, \vec{t}_i),
	\end{equation}
	
	the superior-proposition is defined by
	
	\begin{equation}
		\label{tlo_action_selection}
		\begin{split}
			\text{sup}&(a^*, a, s, i) := \vec{Qt}(s,a^*)_i > \vec{Qt}(s,a)_i \lor\\
			&\big[\vec{Qt}(s,a^*)_i = \vec{Qt}(s,a)_i \land (i=I \lor \text{sup}(a^*, a, s, i+1))\big].
		\end{split}
	\end{equation}

	The TLO policy $\bar{\pi}_\text{TLO}$ then assigns the associated TLO action $a^*$ to each state $s$. A noisy version of $\bar{\pi}_\text{TLO}$ is used during learning for exploration. 
	
	In the original TLO approach \cite{gabor1998multi}, the thresholded form of the Q-function $\vec{Qt}(s,a)$ is learned, by applying value iteration updates to it directly. This potentially leads to massively biased estimations of $\vec{Qt}(s,a)$, especially when TLO is combined with function approximation \cite{li2019urban}. In the modified TLO version named TLQ \cite{vamplew2011empirical}, instead, the vector-valued $\vec{Q}$ function is learned, and $\vec{Qt}$ is calculated based on $\vec{Q}$ during action selection. For learning $\vec{Q}$, the standard Q-learning update is applied independently per objective
	
	\begin{equation}\label{veQ_bellman_update}
		\vec{Q}(s,a)_i\leftarrow \vec{Q}(s,a)_i + \alpha\big[\vec{r}_i + \gamma_i\ \text{max}_{a'\in A}\vec{Q}(s',a')_i-\vec{Q}(s,a)_i\big].
	\end{equation} 
	
	This off-policy update is based on the assumption that the best subsequent action $a'$ is chosen to maximize the current objective $i$ independently from other objectives. In \cite{li2019urban}, Li et al. also propose to learn $\vec{Q}$, but avoid the mentioned problems by taking the TLO action selection into account during the update.% (as in \cite{van2013scalarized}). %This is done by limiting the action set from which $a'$ is chosen within the update.

	\section{gTLO: generalized Thresholded Lexicographic Ordering}\label{method}

	As in other generalized MORL approaches, we aim to learn a Q-function $\mathbfcal{Q}(\vec{s},a,\vec{w},\theta)$ that generalizes over preference parameters $\vec{w}$, where in our case instead of linear scalarization the TLO action selection mechanism is applied. Therefore, in our case, the preference parameters $\vec{w}$ are the TLO threshold values $\vec{t}$, and $\vec{Q}(s,a,\vec{t})_i$ denotes the expected reward $\vec{r}_i$ for reward-component $i$ under the condition, that the agent fulfills the threshold constraints. In contrast to other TLO based algorithms, where either a single policy for a predefined threshold-vector $\vec{t}$ \cite{gabor1998multi, vamplew2011empirical} or for an adaptive $\vec{t}$ \cite{li2019urban} is learned, we learn a generalized Q-function $\vec{Q}(s,a,\vec{t})$ and thereby a generalized form of the TLO policy $\bar{\pi}_\text{TLO}(s,\vec{t})$.
	
	To simplify the following description of \textit{gTLO}, we reformulate the TLO action selection mechanism from Section \ref{TLO} based on sets $\hat{A}_{(\vec{t},i,s)}\subseteq A$ of actions for which the expected future rewards $\vec{Q}(s,a,\vec{t})_j$ are sufficient regarding the thresholds $\vec{t}_j$ up to $i$ for state $s$
	
	\begin{equation}\label{ai_set}
		\hat{A}_{(\vec{t},i,s)}: = \{a\in A \mid \vec{Q}(s,a,\vec{t})_j > \vec{t}_j, \forall j\in [0,i] \}.
	\end{equation}
	
	As shown in detail in appendix \ref{app_equivalence}, the deterministic policy $\bar{\pi}_{\text{TLO}}$, which greedily follows the TLO action selection for a threshold-vector $\vec{t}$, can then be expressed by the conditioned argumentum maximi
		
	\begin{equation}\label{TLO_policy}
		\bar{\pi}_\text{TLO}\leftarrow 
		\begin{cases}
			\text{arg max}_{a\in A} \vec{Q}(s,a,\vec{t})_0, & \text{if }\abs{\hat{A}_{(\vec{t},0,s)}} = 0,\\
			\text{arg max}_{a\in \hat{A}_{(\vec{t},I,s)}} \vec{Q}(s,a,\vec{t})_I, & \text{if } \abs{\hat{A}_{(\vec{t},I,s)}} > 0,\\
			\text{arg max}_{a\in \hat{A}_{(\vec{t},i,s)}} \vec{Q}(s,a,\vec{t})_ {i+1}, & \text{otherwise,}
		\end{cases}
	\end{equation}
	
	where in the third case $i:=\max_{i}: \abs{\hat{A}_{(\vec{t},i,s)}} > 0$. In the following, we specify our idea of a generalized Q-function for TLO and then introduce the deep \textit{gTLO} Network.
	
	\subsection{Generalized TLO Q-Function}
	
	The direct per-objective off-policy Q-learning update as in (\ref{veQ_bellman_update}) implies the assumption that $a'$ is chosen to maximize the current objective independently from other objectives. As shown in (\ref{ai_set}) and (\ref{TLO_policy}), the TLO policy $\bar{\pi}_\text{TLO}(s,\vec{t})$ chooses $a$ from a set of actions $\hat{A}_i$, which itself depends on expected reward values $\vec{Q}(s,a,\vec{t})_j$ for objectives $j<i$. Learning seems to be possible in the single-policy tabular case despite the systematic positive bias resulting from this incorrect assumption as shown in \cite{vamplew2011empirical}. However, in our case of learning a generalized Q-function with function approximation methods, we observe a massive negative impact of this bias on the learning performance. 
	
	In some tabular methods with vector-valued Q-functions the on-policy SARSA update is used to avoid this problem with the off-policy update \cite{sprague2003multiple}. However, value-based deep reinforcement learning in general and generalized MORL in particular heavily rely on historic data for Q-function approximation. In addition to the policy shift, these data may be experienced under outdated preferences. These off-policy data can not be used for on-policy learning. Instead, inspired by \cite{li2019urban}, where a similar update is proposed for single-policy MORL with adaptive thresholds, we define the $\vec{Q}$-function update in a way that considers the TLO action selection mechanism by restricting the set from which the hypothetical follow-up action $a'$ is chosen:
	
	\begin{equation}\label{TLObellman_update}
		\vec{Q}(s,a,\vec{t})_i \leftarrow \vec{Q}(s,a,\vec{t})_i + \\
		\alpha \tau_\text{TLO},
	\end{equation}

	where 
	
	\begin{equation}
		\tau_\text{TLO} := \vec{r}_i + \gamma_i\ \text{max}_{a'\in\tilde{A}_{(\vec{t},i,s')}}\vec{Q}(s',a',\vec{t})_i-\vec{Q}(s,a,\vec{t})_i
	\end{equation}
	
	is a special form of the \textit{temporal difference error}, which is based on the restricted action set

	\begin{equation}\label{TLObellman_update_tildeA}
		\tilde{A}_{(\vec{t},i,s)}:=
		\begin{cases}
			\hat{A}_{(\vec{t},i-1,s)}, & \text{if } \abs{\hat{A}_{(\vec{t},i-1,s)}} > 0,\\
			\{a|a=\bar{\pi}_{\text{TLO}}(s',\vec{t})\}, & \text{else.}
		\end{cases}
	\end{equation}

	In cases where no actions are known that fulfill the threshold conditions up to objective $i-1$, and thus $\hat{A}_{(\vec{t},i-1,s)}=\emptyset$, it is assumed that $a'$ is chosen by following the TLO action selection mechanism. As estimations $\vec{Q}(s,a,\vec{t})_i$ are not used for action selection until $\abs{\hat{A}_{(\vec{t},i-1,s)}} > 0$ (compare (\ref{TLO_policy})), the update in these cases has the purpose to learn the initial expected values until the agent finds a way to fulfill the constraints up to $i$ for state $s$ and action $a$.
	
	\subsection{Deep gTLO Network}
	\label{deep_gtlo_network}
	
	\begin{figure}
		\centering
		\includegraphics[width=0.7\linewidth]{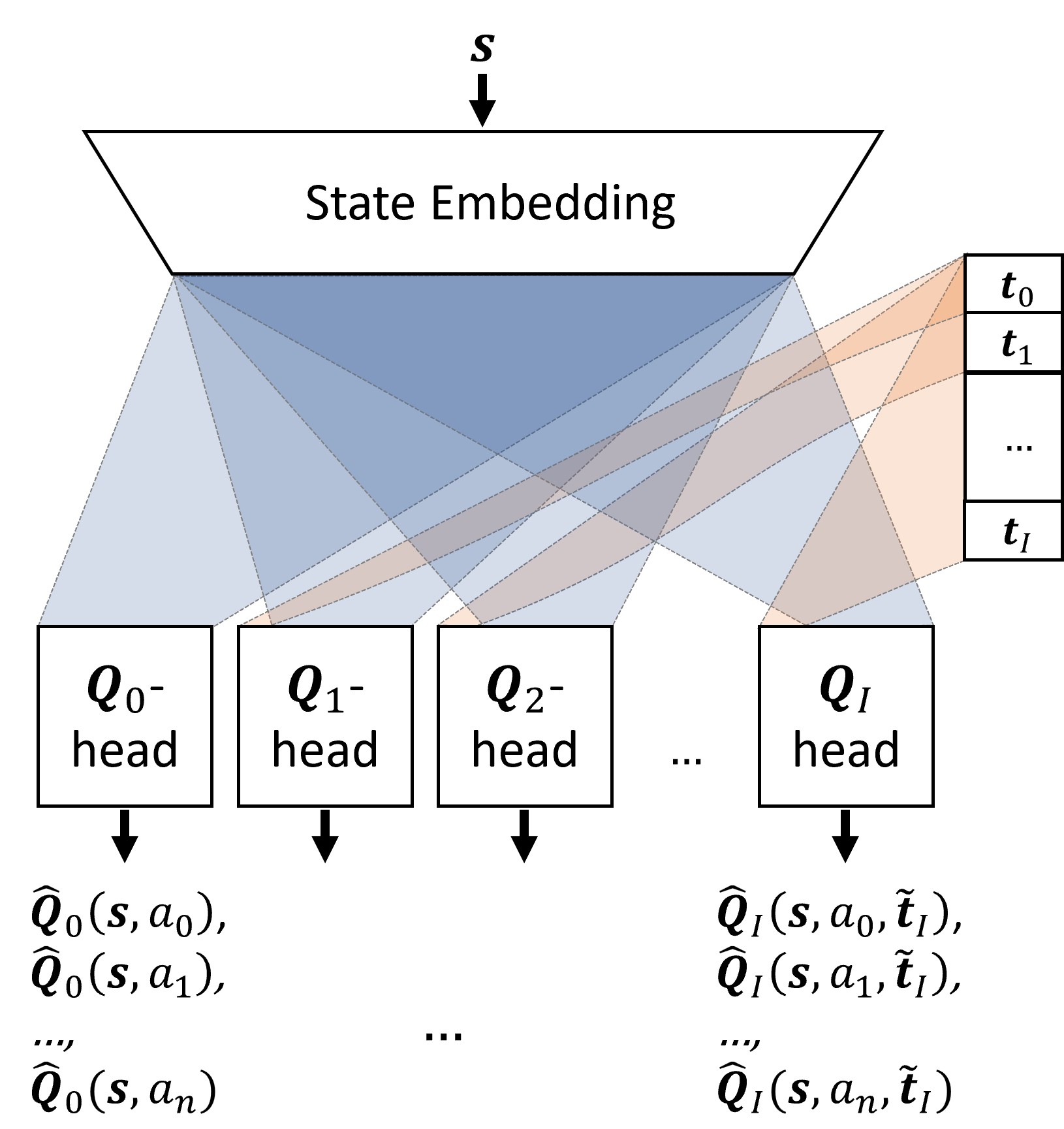}
		\caption{General model architecture of the generalizedTLO Network}
		\label{fig:generalizedtlonetworkarchitecture}
	\end{figure}
	Generalized MORL methods use function approximation to generalize across different preference parameters. In our case, preference parameters are the TLO thresholds $\vec{t}$. We use deep neural networks to approximate the vector-valued multi-objective Q-function $\mathbfcal{Q}(s,a,\hat{\vec{t}},\theta)\approx\vec{Q}(s',a',\hat{\vec{t}})$. Vanilla DQN \cite{mnih2015human} serves as the base algorithm of our \textit{gTLO} implementation. However, we want to emphasize that \textit{gTLO} is compatible with extended DQN versions and other value-based approximate or deep reinforcement learning algorithms.%also compatible to other value based methods such as DDPG \cite{lillicrap2015continuous} in the case of continuous actions. %\textbf{its in fact an hybrid of both cases in the linear section, in the linear case, vector valued prediction does not have to be conditioned}
		
	The extension of a base algorithm for \textit{gTLO} learning is straightforward: The TLO action selection (\ref{TLO_policy}) replaces the default greedy policy and for learning, we use an $\epsilon$-greedy version of it. The multi-objective Q-function is approximated by a multi-headed deep neural network $\mathbfcal{Q}$ as depicted in Fig. \ref{fig:generalizedtlonetworkarchitecture}. The loss-function for $\mathbfcal{Q}$ is derived from the \textit{gTLO} update defined in (\ref{TLObellman_update}) to (\ref{TLObellman_update_tildeA}).
	
	The \textit{gTLO} network consists of state embedding layers followed by per-objective heads $\mathbfcal{Q}_i$. The state embedding layers embed the state $s$ and usually consist either of convolutional layers followed by a flattening layer in the case of image states or fully-connected layers in the case of vector-valued state descriptions. Objective heads $\mathbfcal{Q}_i$ consist of fully-connected layers and map from state embeddings to $\vec{Q}_i$ estimations. While the state embedding is independent of thresholds $\vec{t}$, the $\mathbfcal{Q}_i$ heads are conditioned on $\vec{t}$. When revisiting the definitions of the action sets $\hat{A}_{(\vec{t},i,s)}$ (\ref{ai_set}) and $\tilde{A}_{(\vec{t},i,s)}$ (\ref{TLObellman_update_tildeA}) underlying the TLO action selection mechanism (\ref{TLO_policy}) and the Q-function update (\ref{TLObellman_update}), one sees that in both cases only thresholds $\vec{t}_j$ up to a certain degree are taken into account. In particular, the update and therefore the approximation $\mathbfcal{Q}_i$ depends on thresholds $\vec{t}_j$ with $j<i$. Therefore, it is sufficient to take the vector $\hat{\vec{t}}_i = (\vec{t}_0, ..., \vec{t}_{i-1})^\top$ into account when learning $\mathbfcal{Q}_i$. Expected values for the first objective $\mathbfcal{Q}_0$ can be learned without considering $\vec{t}$ at all.
	
	The loss for training the \textit{gTLO} network is derived from the TLO bellman update (\ref{TLObellman_update}). The \textit{gTLO} training target for the experience tuple $(s,a,s',\vec{r})$ and the objective with index $i$ corresponds to the standard DQN target (\ref{dqn_target}):
	\begin{figure*}
		\centering
		\includegraphics[width=0.7\linewidth]{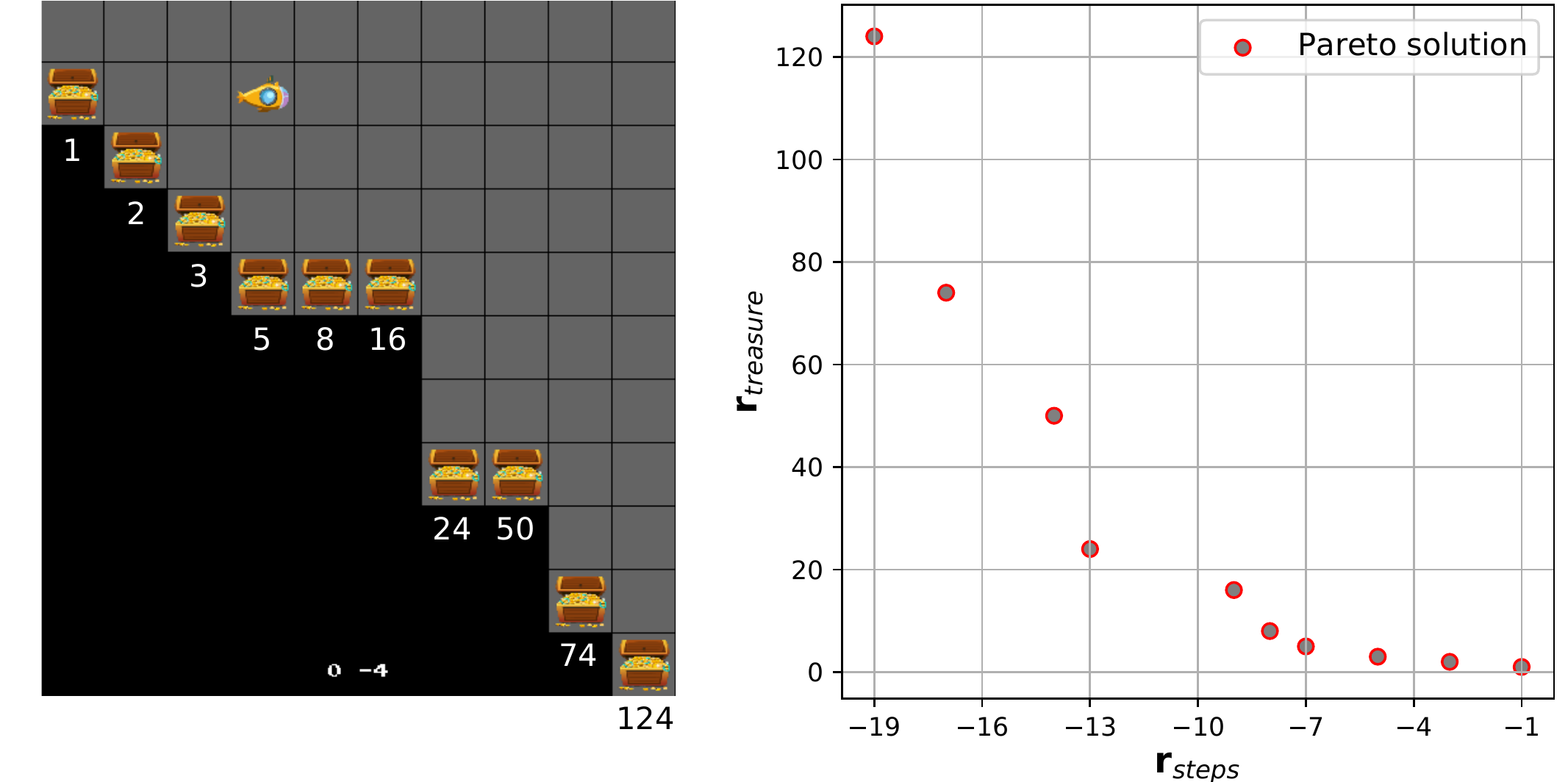}
		\caption{Deep-sea treasure environment. Left: Screenshot of the DST environment used during evaluation with annotated per treasure rewards. Right: Pareto solutions of the environment in the space of returns.}
		\label{fig:dst}
	\end{figure*}
	\begin{equation}\label{gTLO_target}
		y_{\text{gTLO},i} := \vec{r}_i + \gamma\ \text{max}_{a'\in\tilde{A}_{(\vec{t},i,s')}}\mathbfcal{Q}(s',a',\hat{\vec{t}},\theta^-)_i.
	\end{equation}
	
	The per-sample loss is then the sum of per objective losses
	
	\begin{equation}\label{gTLO_loss}
		\mathcal{L}(\theta) := \sum_{i\in[1,I]}\mathcal{H}\big(y_{\text{gTLO},i}-\mathbfcal{Q}(s,a,\hat{\vec{t}},\theta)_i,\delta\big),
	\end{equation}
	
	where $\mathcal{H}$ is the Huber loss with $\delta = 1$, which was shown to stabilize DQN \cite{mnih2015human} training.
	
	\section{Experiments}\label{experiments}
	
	\subsection{Deep-Sea Treasure Environment}\label{dst}

	Deep-sea treasure (DST) \cite{vamplew2008limitations} is a simple benchmark grid world that is often used to investigate multi-objective reinforcement learning algorithms. In DST, the agent controls a submarine, where the four available actions per step are to move one cell left, right, up and down. When the resulting cell would exceed a grid world border, the submarine's position remains on the current cell. Each episode starts at the upper left cell and ends when the submarine lands on special terminal cells with assigned treasures. The objective of the agent is to maximize the value of treasures collected while using as few steps as possible. The treasure value is reflected by a positive treasure-reward signal $\vec{r}_\text{treasure}$, which is emitted when a treasure is reached. An efficiency-reward signal $\vec{r}_\text{steps}=-1$ is emitted in each step.
	
	One of the key features of DST is the possibility to easily define environments with various Pareto front shapes. Our evaluation is based on an image version of the original DST environment \cite{vamplew2008limitations}, which is shown in Fig. \ref{fig:dst} (left). It consists of $11\times 10$ cells, with one treasure per column. Treasure values and positions are chosen such that the Pareto front, visualized in Fig \ref{fig:dst} (right), is non-convex.
	
	As originally proposed in \cite{mossalam2016multi}, we use an image version of DST. In the original DST environment, the state is encoded by a one-hot-vector of the submarine's position for investigating tabular MORL methods. In the image version, the state is represented by a down-sampled $84\times 84$ grey-scale image of the current screen output. Our implementation of the environment is based on the \textit{Fruit API} DST implementation \cite{nguyen2020review}. In our implementation, each DST episode ends when a treasure is reached or with a treasure reward of $\vec{r}_\text{treasure}=0$ when 50 time steps have passed.
	
	\subsection{Deep Drawing MORL Environment}
	\label{deepdrawing_env}
	% dornheim2018multiobjective
	\begin{figure*}
		\centering
		\includegraphics[width=0.9\linewidth]{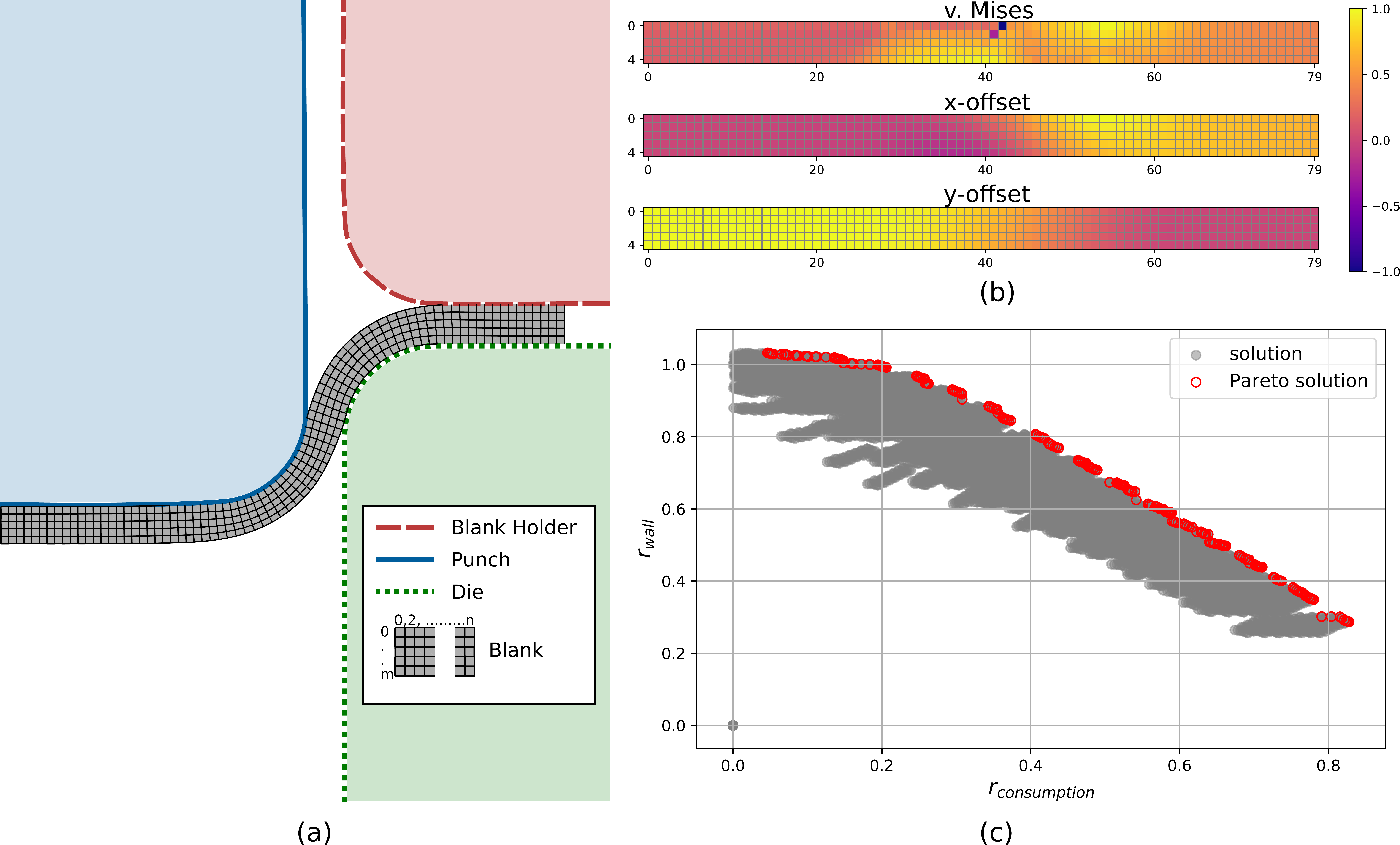}
		\caption{Deep drawing MORL environment. (a) Components of the rotationally symmetric deep drawing simulation model, where the blank (grey mesh) is drawn by the punch (blue) into the hollow inner part of a die (green), and the blank holder regulates the material flow. (b) Spatially resolved parameters of the blank, as used to describe the process state $\vec{s}$. (c) Solutions of the MODP in the space of per-objective returns with Pareto front solutions marked in red.}
		\label{fig:deepdrawing}
	\end{figure*}
	To investigate the application of \textit{gTLO} to real-world problems, we use a multi-objective variant of the deep drawing optimal control environment from \cite{dornheim2020model}. A profile view of the deep drawing process is depicted in Fig. \ref{fig:deepdrawing}(a). It is a steel deformation process, where a stamp draws a metal blank into a so-called die to produce a hollow form. Blank holders regulate the material flow. The quality of the process outcome depends to a large extent on the blank holder forces \cite{singh2015study}. The online optimization of time-varying blank holder forces regarding the process outcome is the objective of the deep drawing environment. In contrast to the deep-sea treasure environment, the deep drawing process is subject to stochastic conditions in form of a varying friction coefficient. Also, as Fig. \ref{fig:deepdrawing}(c) shows, the amount of policies on the Pareto front is vastly superior when compared to the deep-sea treasure environment.

	The deep drawing environment is based on an interactive numerical simulation of the deep drawing process. Each simulated process execution is reflected by an episode of five time steps, in which the process simulation is interrupted and the agent determines the next blank holder force from a discrete action set $A=\{2 \text{kN}, 4 \text{kN}, 6 \text{kN}, 8 \text{kN}, 10 \text{kN}, 12 \text{kN}, 14 \text{kN}\}$. At the end of each process execution, a reward is signaled based on a quality assessment of the deep-drawn workpiece. In the multi-objective variant, two conflicting objectives are (a) to reach a leveled wall thickness in the resulting cup and (b) to minimize the material consumption of the process. These objectives are reflected by a reward vector $\vec{r}\in\mathbb{R}^2$. Both reward signals are aggregated from the simulation results as specified in \cite{dornheim2020model}. Instead of the three scalar sensor values which are used in \cite{dornheim2020model} to describe the process state, we assume the observability of stress and geometry data, which we use as state description $\vec{s}\in \mathbb{R}^{5\times 80\times 3}$. Stress information is given in the form of the von Mises stress in each of the $5\times 80$ elements of the mesh. Geometry information is given by the offset in x- and y-direction regarding the initial positions of the $5\times 80$ blank elements. An exemplary state is visualized in Fig. \ref{fig:deepdrawing}(b). As in the original environment \cite{dornheim2020model}, The friction coefficient $\mu$ is drawn randomly and independently per episode from a beta distribution with $(\alpha=1.75, \beta=5)$, which is rescaled to the range $[0, 0.14]$ and discretized into 10 equally sized bins. Discretization is performed with the purpose of enabling the reuse of previous results of the expensive numerical simulation. Due to the varying process conditions, the optimal policy depends on the friction coefficient. As an exemplary excerpt of the solution space, the set of all solutions with the Pareto front marked in red is visualized in Fig. \ref{fig:deepdrawing}(c) for $\mu=0.028$, which is the mode of the discretized friction distribution. For a detailed description of the environment and the underlying FEM simulation, we refer to \cite{dornheim2020model}. 
	
	\subsection{Evaluation Procedure and Metrics}\label{metrics}
\begin{figure}
		\centering
		\includegraphics[width=0.5\linewidth]{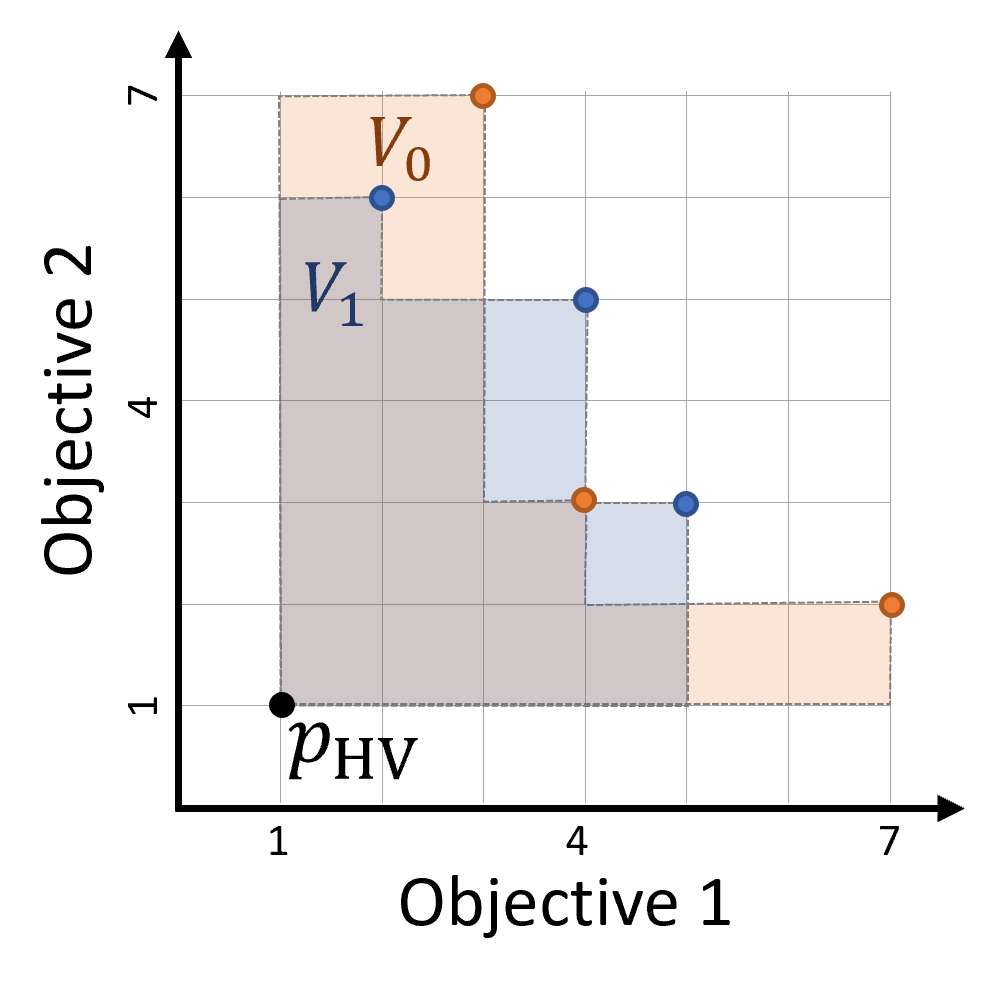}
		\caption{Hypervolume regions $V_0$ and $V_1$ (orange and blue shaded areas) for two solution sets (orange and blue scatters) and the reference point $p_\text{HV}$.}
		\label{fig:hypervolume}
	\end{figure}

	Specific metrics are used to evaluate multi-objective multi-policy methods. The \textit{hypervolume metric} \cite{zitzler2003performance} is broadly used in multi-objective optimization to evaluate sets of found solutions. It is defined for a set of solutions and a reference point $p_\text{HV}$ as the volume of the area between the solutions and $p_\text{HV}$ and strictly increases with regard to the Pareto dominance relation \cite{van2013hypervolume}. For an exemplary two-objective problem, hypervolumes $V_0=17$ and $V_1=15$ for two exemplary solution-sets, each consisting of three solutions, are depicted in Fig. \ref{fig:hypervolume}, where each hypervolume equals the allocated shaded area. As proposed in \cite{vamplew2011empirical} for the case of multi-policy MORL, we use the hypervolume metric as an offline metric, which is periodically calculated during learning. To achieve this, the learning process is interrupted every few learning episodes to perform an evaluation phase. Within the evaluation phase, a solution-set $\mathcal{S}$ is sampled by systematically varying the preferences $\vec{w}$ and acting greedily regarding the current state of the $Q$-function approximations. 
	
	If the environment is deterministic and the set of Pareto front solutions $\mathcal{P}$ is finite and known, as it is the case in the deep-sea treasure environment, metrics can be calculated directly based on the solutions found $\mathcal{S}$ and the Pareto solutions $\mathcal{P}$ \cite{yang2019generalized}. For further insight on the deep-sea treasure results, we report the precision ($\text{precision}=\abs{\mathcal{S}\cap \mathcal{P}} / \abs{\mathcal{S}}$), the recall ($\text{recall}=\abs{\mathcal{S}\cap \mathcal{P}} / \abs{\mathcal{P}}$) and the F1 score
	
	\begin{equation}\label{F1}
		F_1 := 2\times\dfrac{\text{precision}\times \text{recall}}{\text{precision}+\text{recall}},
	\end{equation}
	
	as suggested in \cite{yang2019generalized}.
	
	For deterministic environments, sampling the policy for a single episode per preference $\vec{w}$ is sufficient to calculate the exact offline metrics. In the case of a stochastic environment, however, repeated independent executions are necessary to calculate the statistics of the respective metric. As described in Section \ref{deepdrawing_env} for the deep drawing environment, the probability distribution of the stochastic friction coefficient $\mu$, is discrete and although it is not known to the agent, it can be manually set during the evaluation phase. We take advantage of this by systematically varying the friction coefficient during the evaluation phase and running one evaluation episode for each combination of friction and preference. On this basis, the expected values of evaluation metrics can be calculated also for the deep-drawing environment.
	
	\subsection{Methods and Implementation Details}
	\label{impl_details}
	
	Methods we considered in the evaluation studies are:
	
	\begin{itemize}
		\item \textit{gTLO}: our generalized TLO approach as described,
		\item \textit{gLinear}: generalized MORL based on linear scalarization and a generalized Deep Q-Network 
		\item \textit{dTLQ (outer loop)}: The deep TLQ approach from \cite{nguyen2020multi}.
	\end{itemize}
	
	To isolate the effect of generalization, we further consider an outer loop variant of our approach which we call \textit{outer loop gTLO}. Using the popular non-linear MORL benchmark deep-sea treasure further allows us to relate our results to previously published results.
	
	The method \textit{gLinear} is based on a scalar, generalized deep Q-network, which is trained to estimate the \textit{expected scalarized return}. It is representative of other linear methods, including current generalized methods, which may converge faster but are also limited to the \textit{convex coverage set}. 
	
	 \textit{gTLO}, its variants, and \textit{gLinear} are implemented based on the DQN implementation of the \textit{stable-baselines} framework \cite{stable-baselines}. The network architecture for the \textit{gTLO} Network as used for the DST experiments is as follows:
	\begin{itemize}
		\item The \textit{state embedding module} $\vec{s}\rightarrow\tilde{\vec{x}}$ is a slightly altered version of the DQN architecture \cite{mnih2015human} and consists of three convolution layers with $32, 64$, and $64$ filters, a filter-size of $8,4$, and $3$ and a stride of $4,2,$ and $1$ followed by a fully-connected layer of $256$ neurons.
		\item The $\mathbfcal{Q}_0$-head $\tilde{\vec{x}}\rightarrow(\mathbfcal{Q}(\vec{s},a_0)_0,\mathbfcal{Q}_0(\vec{s},a_1),...,\mathbfcal{Q}_0(\vec{s},a_n))$ consists of a single fully-connected layer of $128$ neurons, followed by the linear output Layer.
		\item The $\mathbfcal{Q}_1$-head $(\tilde{\vec{x}}, \vec{t}_0)\rightarrow(\mathbfcal{Q}_1(\vec{s},a_0),\mathbfcal{Q}_1(\vec{s},a_1),...,\mathbfcal{Q}_1(\vec{s},a_n))$ consists of two fully-connected layers of $128$ and $64$ neurons, followed by the linear output Layer.
		\item The rectified linear unit (ReLU) is used as activation for non-linear layers.
	\end{itemize}
	
	For the deep drawing experiments, we decreased the capacity of the state embedding module, by reducing it to two convolutional layers with $32$ and $64$ filters. Due to the wide format of the input image (see Fig. \ref{fig:deepdrawing}(b)), filter sizes and strides are reduced to $3$ and $1$ respectively. 
	
	The scalar generalized networks used for the \textit{gLinear} implementation are equal to the \textit{gTLO} state embedding modules followed by a linear output layer.
	
	%\textbf{hyper-parameter global}
	In general, we train all networks with 8 mini-batches per time step and do not limit the replay-buffer size. 
	We evaluate \textit{dTLQ} based on the implementation as part of the FruitAPI framework of the \textit{dTLQ} author \cite{nguyen2020multi}, which we extended to enable multiple mini-batches per time step.
	
	%\textbf{hyper-parameter DST}
	For DST, one experiment run consists of $250,000$ training time steps with an evaluation phase every $1,000$ time steps. For evaluating \textit{gTLO}, the preferences $\vec{t}_0$ are drawn independently per training episode from a set of $100$ equidistant values from the interval $[0.5, 100]$. Both environments are two-objective environments, thereby formally, $\vec{t}_1=\infty$ in both cases. Similarly, for evaluationg \textit{gLinear}, weights $\vec{w}$ are drawn per episode from a set of vectors $\vec{w} = \phi(0, 1)^\top+(1-\phi)(1, 0)^\top$ for $100$ equidistant values $\phi\in[0,1]$. 
	For the outer loop algorithms (\textit{dTLQ} and \textit{outer loop gTLO}), preferences $\vec{t}_0$ are drawn from a set of ten preferences, that have been precalculated based on the knowledge of the Pareto front solutions following \cite{nguyen2020multi}: for each pair of each consecutive treasure-values $\vec{r}_\text{treasure}$, $\vec{r}_\text{treasure}'$, including the special case $\vec{r}_\text{treasure}=0$, a preference $\vec{t}_0=(\vec{r}_\text{treasure}+\vec{r}_\text{treasure}')/2$ is added. The reference point for the hypervolume metric is chosen as $p_\text{HV}=(0,-25)$ for DST.
	
	\begin{figure*}[! h]
		\centering
		\subfloat[Hypervolume over training time steps per method]{\includegraphics[width=0.45\textwidth]{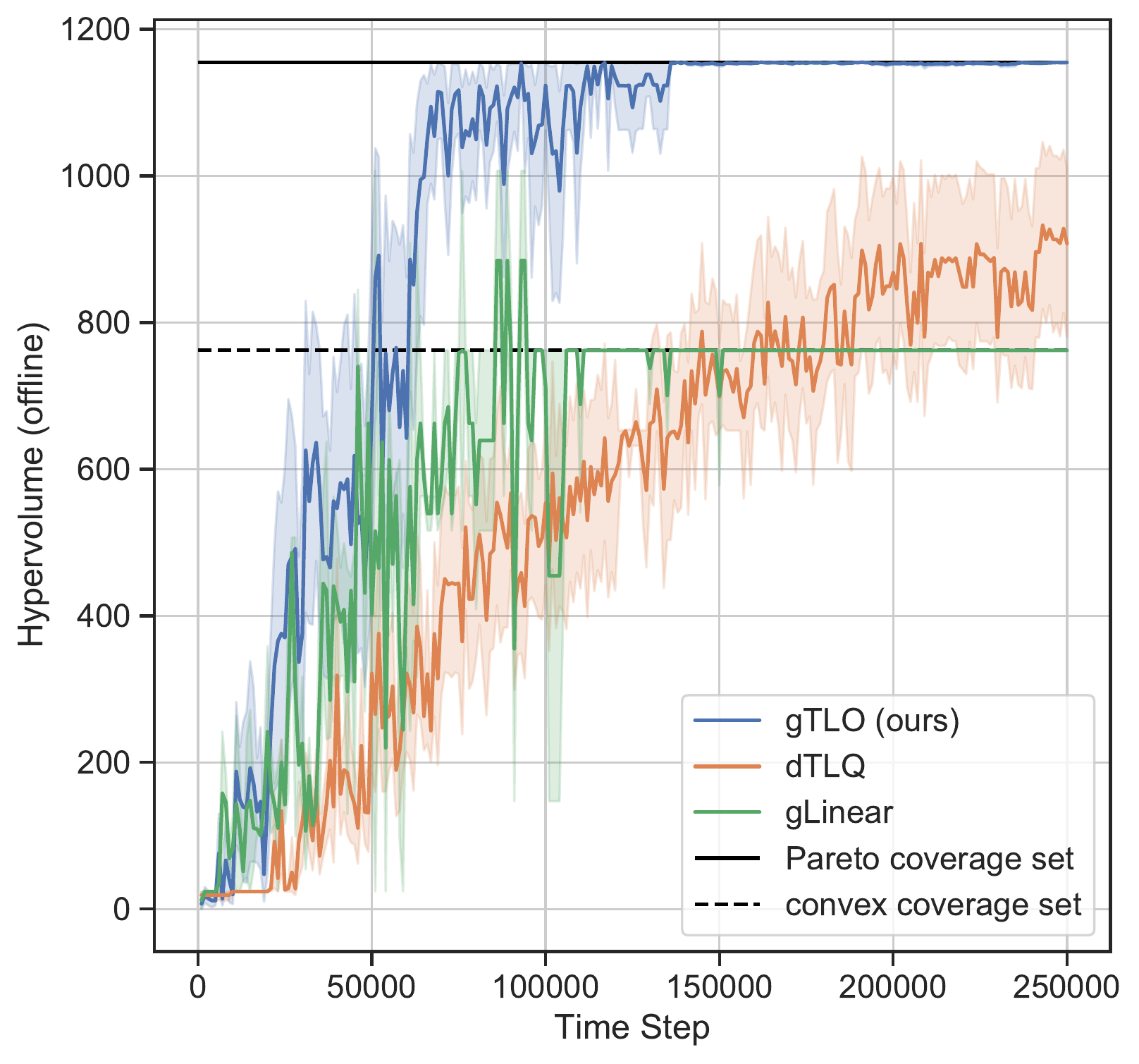}}
		\subfloat[Frequency of optimal solutions per method]{\includegraphics[width=0.45\textwidth]{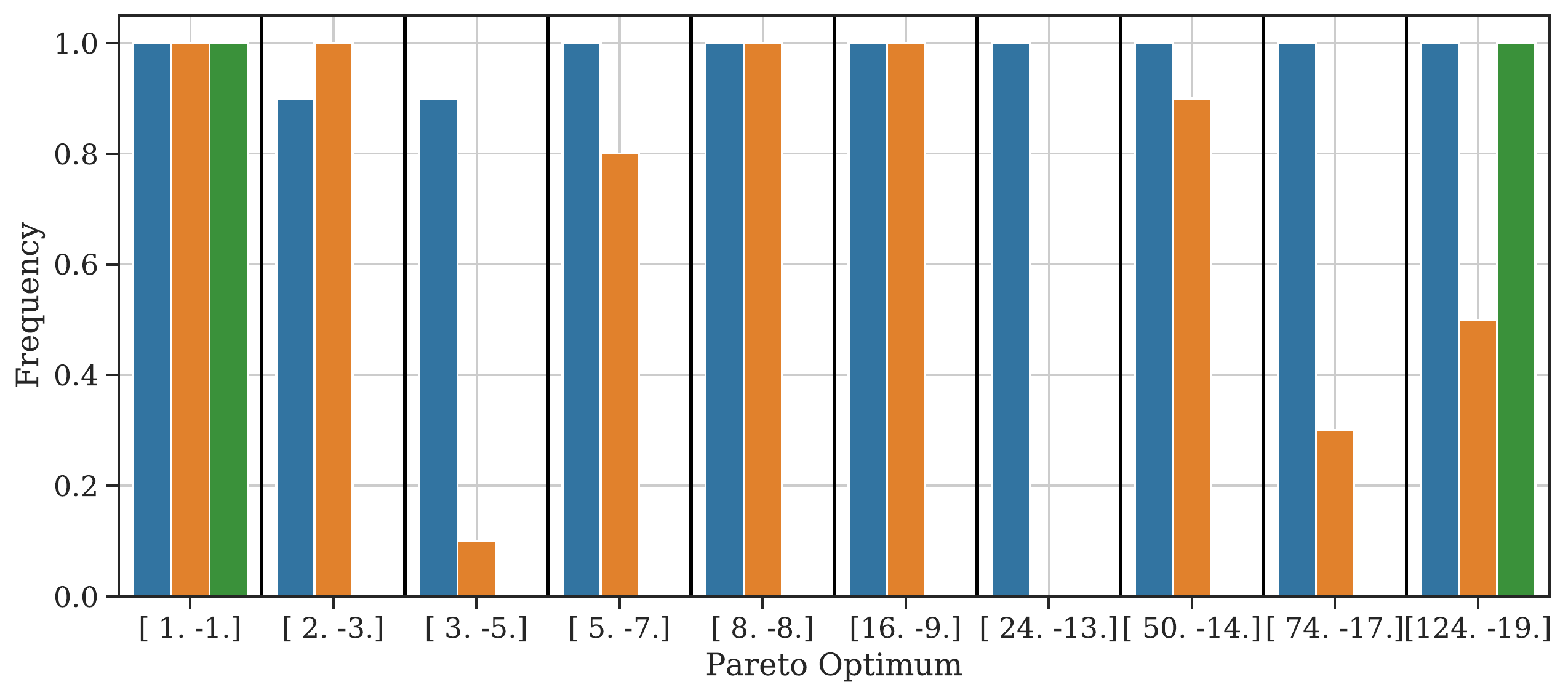}}\\
		\subfloat[Precision over training time steps per method]{\includegraphics[width=0.45\textwidth]{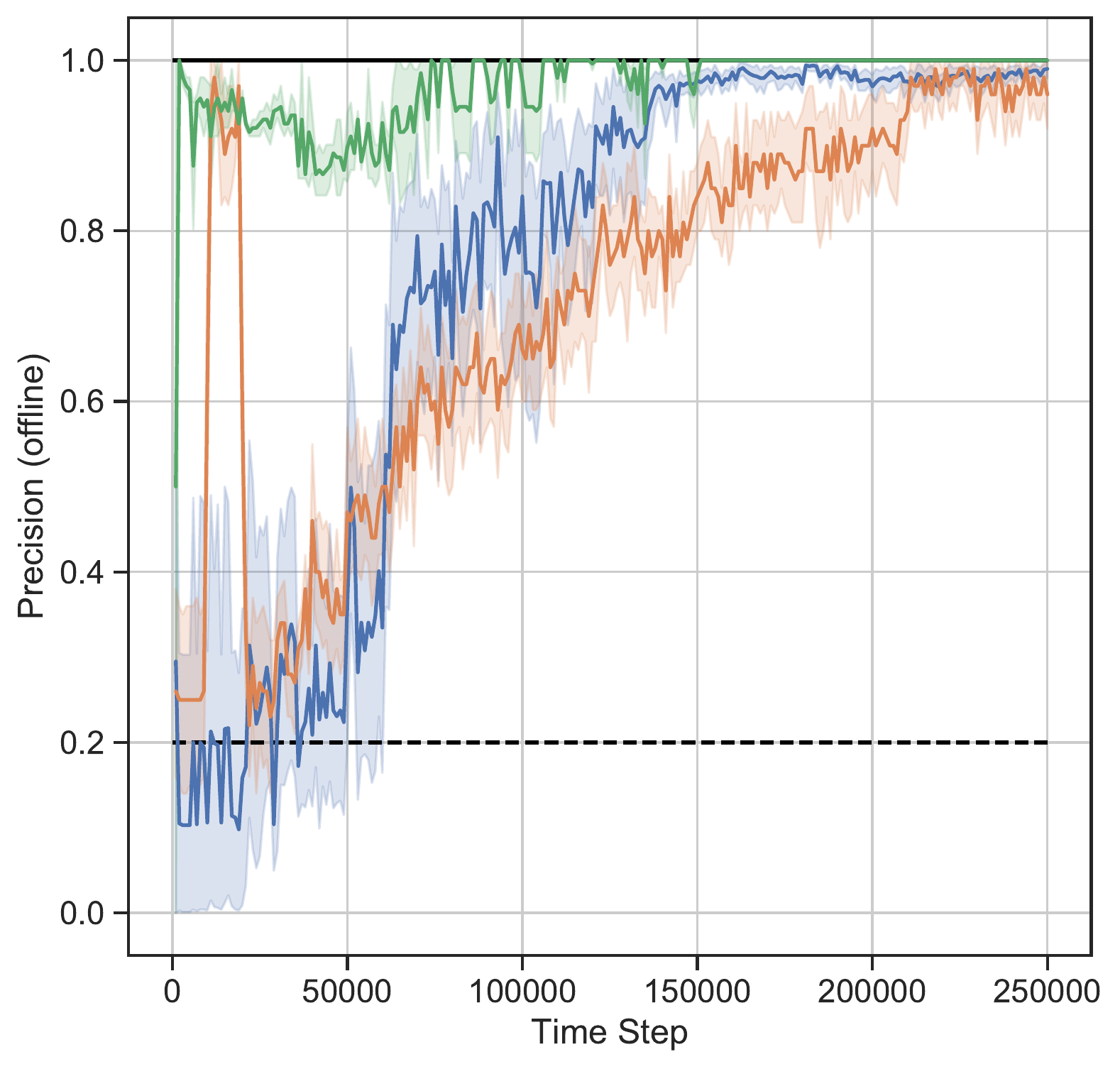}}
		\subfloat[Recall over training time steps per method]{\includegraphics[width=0.45\textwidth]{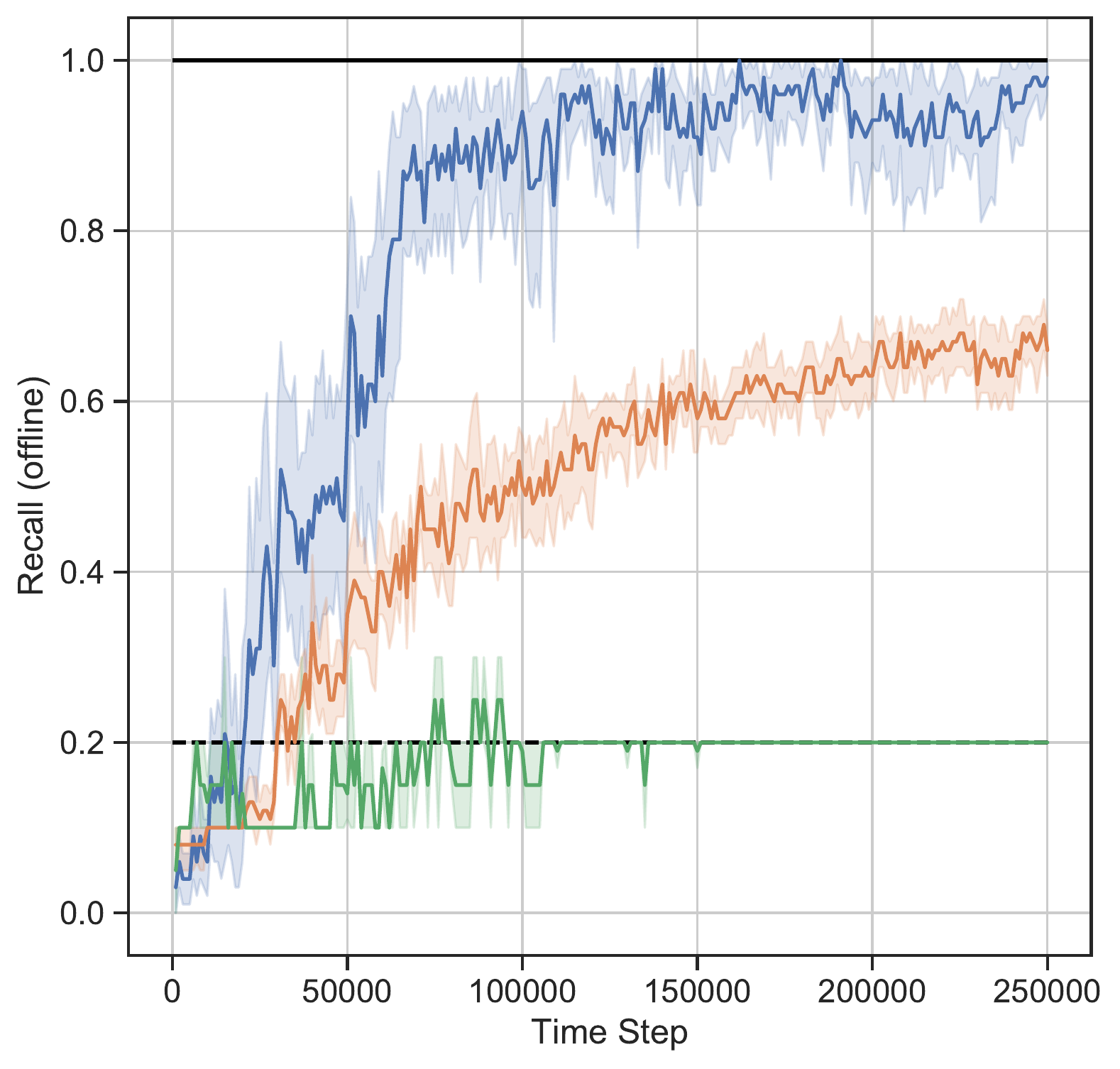}}
		\caption{Deep-sea treasure results. (a) Hypervolume metric, (c) precision and (d) recall over time steps per approach. Frequency of determination of each Pareto front solution at the end of learning per approach (b). The hypervolume metric, precision and recall are calculated offline every $1,000$ time steps. The mean and the $0.95$-confidence interval are plotted per metric and approach. The hypervolume reference is defined as $p_\text{HV}=(0,-25)$.}
		\label{fig:result_dst}
	\end{figure*}
	%\textbf{DST / gTLO}
	%\textbf{DST / Linear}
	%\textbf{DST / outer loop}
	Further DQN parameters for \textit{gTLO} are chosen as follows: the reward is undiscounted ($\gamma=1$), the target network update frequency used is $5,000$ and the warm-up phase lasts $1,000$ time steps. For \textit{gLinear}, the discount factor is $\gamma=0.9$ and the prioritized experience replay is activated with its default paramters ($\alpha_\text{PER}=0.6, \beta_\text{PER}=0.4$). 
	
	%\textbf{hyper-parameter DD}
	Each experiment run on the deep drawing environment consists of $12,500$ training time steps, where the evaluation phase is conducted every $1,250$ time steps. The rewards are scaled to the range $[0,1]$ based on empirical data and then clipped to ensure $\vec{r}_i\in\mathbb{R}^2_+$. The TLO threshold $\vec{t}_0$ is drawn per episode from a set of $n$ equidistant values from the interval $[0, 1.2]$, where for the inner loop (generalized) MORL methods $n=100$ and for \textit{outer loop gTLO} $n=10$. The hypervolume reference point is set to $p_\text{HV}=(-0.1,-0.1)$.
	%reward_terms = ["rt_feeding", "rt_thickness"]
	The target networks are updated every $2,505$ steps. Other DQN parameters and the preference vectors for \textit{gLinear} are chosen as for the DST experiments.
	
	Code and configurations for reproducing the reported results are published as stated in Section \ref{code}.
		
	\section{Results}\label{results}
	\subsection{Deep-Sea Treasure}\label{results_dst}
		
	\begin{table*}[h]
	\begin{center}
		\begin{minipage}{330pt}
			\caption{Results on deep-sea treasure after $250,000$ training steps}\label{table:dst_results}%
			\begin{tabular}{@{}l|l|l|l|l@{}}
				%\toprule
				Approach & HVT$(0,-25)$ & Precision & Recall & F1 \\
				\hline
				\hline
				gTLO (ours) & $\textbf{1154.6}\pm \textbf{0.8}$ & $0.99\pm 0.015$ & $\textbf{0.98}\pm \textbf{0.04}$ & $\textbf{0.985}\pm \textbf{0.022}$ \\
				Linear Scalarization & $762.0\pm 0.0$ & $\textbf{1.0}\pm \textbf{0.0}$ & $0.2\pm 0.0$ & $0.334\pm 0.0$  \\
				\hline
				dTLQ (outer loop) \footnotemark[1] & $907.7\pm 190.61$ & $0.96\pm 0.066$ & $0.66\pm 0.049$ & $0.78\pm 0.044$ \\
				\textit{outer loop gTLO} \footnotemark[2] & $1150\pm 14.34$ & $0.98\pm 0.04$ & $\textbf{0.98}\pm \textbf{0.04}$ & $0.98\pm 0.04$ \\
				%gTLO (ours) EXT\footnotemark[3] & $\textbf{0.99}\pm \textbf{0.03}$ & $\textbf{1154.9}\pm\textbf{0.3}$ & $0.99\pm 0.03$ & $\textbf{0.99}\pm \textbf{0.03}$ \\
				\hline
			\end{tabular}
			\footnotetext[1]{As implemented in the FruitAPI Framework \cite{nguyen2020multi}.}
			\footnotetext[2]{Merged Single-Policy MORL with \textit{gTLO} Networks and TLQ Bellman update.}
			%\footnotetext[3]{gTLO approach with 10 exact configurations, using the knowledge about the Pareto-front.}
		\end{minipage}
	\end{center}
	\end{table*}
	As described in Section \ref{dst}, an image version of the original DST environment is used to evaluate the MORL methods specified in Section \ref{impl_details}. The preference is sampled independently per episode from an equal distribution. 
	%The goal is to determine the dominating policies, i.e. to maximize the hypervolume and the other metrics specified in \ref{metrics}. 
	Per method, ten independent runs of $250,000$ training time steps per run are executed. Every $1,000$ time steps an evaluation phase is carried out as described in Section \ref{metrics}. Evaluation results are visualized in Fig. \ref{fig:result_dst} in form of the mean and $0.95$-confidence interval per evaluation phase of the hypervolume metric in Fig. \ref{fig:result_dst}(a), the precision in Fig. \ref{fig:result_dst}(c), and the recall in Fig. \ref{fig:result_dst}(d). Known metric values for the \textit{Pareto coverage set} and the \textit{convex coverage set} are drawn as a black line and dashed black line respectively. In addition, we show the frequency in which each solution from the Pareto front is found within the last evaluation phase of each run per method in Fig. \ref{fig:result_dst}(b). The mean and standard deviation of each metric at the end of the $250,000$ time steps are listed in Table \ref{table:dst_results} for the three methods and the outer loop variant of \textit{gTLO}.

	As expected and repeatedly shown in other publications (e.g. \cite{vamplew2011empirical, van2013hypervolume}) for linear MORL methods, \textit{gLinear} converges to the two extreme solutions which lie on the convex hull of the non-convex Pareto front (compare Fig. \ref{fig:dst}). \textit{gTLO} manages to repeatedly identify all Pareto front solutions except for rare occurrences (as shown in Fig. \ref{fig:result_dst}(b), the solutions with $(2,-3)$ and $(3,-5)$ are not identified in one of the ten runs). 
	The outer loop approach \textit{dTLQ} also covers the whole range of solutions but has problems identifying some individual solutions (at $(3,-5)$ and $(24,-13)$) and rarely identifies the two solutions of highest treasure value. Therefore, \textit{gTLO} clearly outperforms \textit{dTLQ} regarding the recall and the hypervolume metric. While in the evaluation runs \textit{gLinear} never wrongly determines a dominated policy at the end of learning and achieve a precision-value of $1.0$, \textit{gTLO} in rare cases returns policies that are not part of the Pareto front, and thus reaches a mean precision of $0.99$.
	
	To investigate the effect of generalization on the sample efficiency of our method, we compare the convergence speed of \textit{gTLO} and the outer loop variant of \textit{gTLO}. The mean amount of steps needed by the outer loop variant until the full Pareto front is found for the first time ($141,555.55\pm 20,600.97$ steps) is more than double the amount of steps needed by default, inner loop, \textit{gTLO} ($61,000\pm 20,600.97$ steps). In accordance with already published results for linear generalized methods \cite{castelletti2011multi, abels2019dynamic}, this shows a positive effect of generalization regarding the sample efficiency also in the non-linear case.
	
	As the deep-sea treasure environment and variants are used for over a decade for evaluating linear and non-linear methods, it is possible to put our results into relation. While often the classical vector version of DST is used, where the state is encoded as one-hot-vector of the submarine's position \cite{reymond2019pareto, yang2019generalized, van2014multi, vamplew2011empirical, ruiz2017temporal}, some previous deep MORL evaluations are also based on image versions \cite{mossalam2016multi, abels2019dynamic, nguyen2020multi}. Evaluations of linear methods \cite{mossalam2016multi, abels2019dynamic, yang2019generalized} are based on convex DST variants. They would converge to the same solution as \textit{gLinear} when applied to the original non-convex DST. The only method that was evaluated on a non-convex image version of DST is \textit{dTLQ} \cite{nguyen2020multi}, which is shown to solve simplified non-convex DST environments with 3 and 5 treasures after about $300,000$ and $1,000,000$ total steps respectively.
	
	When it comes to exact, tabular non-linear MORL methods, there are some \cite{van2014multi, vamplew2011empirical, ruiz2017temporal} which completely solve the vector-variant of DST. The steps required until convergence are between $75,000$ for TLQ \cite{vamplew2011empirical}, which requires knowledge about the Pareto front, and $1,250,000$ steps for MPQ-learning \cite{ruiz2017temporal}. Other non-linear tabular methods don't completely solve the problem: Q-learning with a Chebyshev scalarization function converges to a hypervolume of $938.29$ \cite{van2013scalarized} and a hypervolume-based inner loop approach reaches $1040.24$ \cite{van2013hypervolume} of the maximum hypervolume of $1,155$.
	
	\textit{Pareto DQN} \cite{reymond2019pareto}, the only other inner loop non-linear deep MORL method we identified is evaluated on the one-hot-vector version of DST. In contrast to our work, in \cite{reymond2019pareto}, details about the Pareto front estimation from the perspective of the first state $s_0$ are reported instead of agent performance metrics. This prevents a direct comparison of results. Although the general course of the Pareto front is approximated by Pareto DQN, the distribution of the estimated Pareto front solutions differ greatly from the true Pareto front. Furthermore, the estimation quality of PDQN highly depends on the complexity of the state space \cite{reymond2019pareto}. %While the true Pareto front consists of $7$ solutions with $r_\text{treasure}\leq 24$, the estimated Pareto front 
	
	\subsection{Deep Drawing}
	\label{results_deepdrawing}
	
	We evaluate \textit{gTLO}, \textit{gLinear}, and the outer loop variant of \textit{gTLO} on the multi-objective deep drawing environment as it is described in Section \ref{deepdrawing_env}. The evaluation is based on ten independent learning runs per method. As the Pareto front is not known for this environment, ten preferences $\vec{t}_0$ used for \textit{outer loop gTLO} are equally spaced between $0.1$ and $1$.
	%Each run consists of $2,500$ episodes and an evaluation phase is accomplished every $250$ learning episodes. %For quantitative comparison, the expected hypervolume is calculated per evaluation phase as described in \ref{metrics}. \textbf{korrektur lesezeichen}
	
	\begin{figure}
		\centering
		\includegraphics[width=0.9\linewidth]{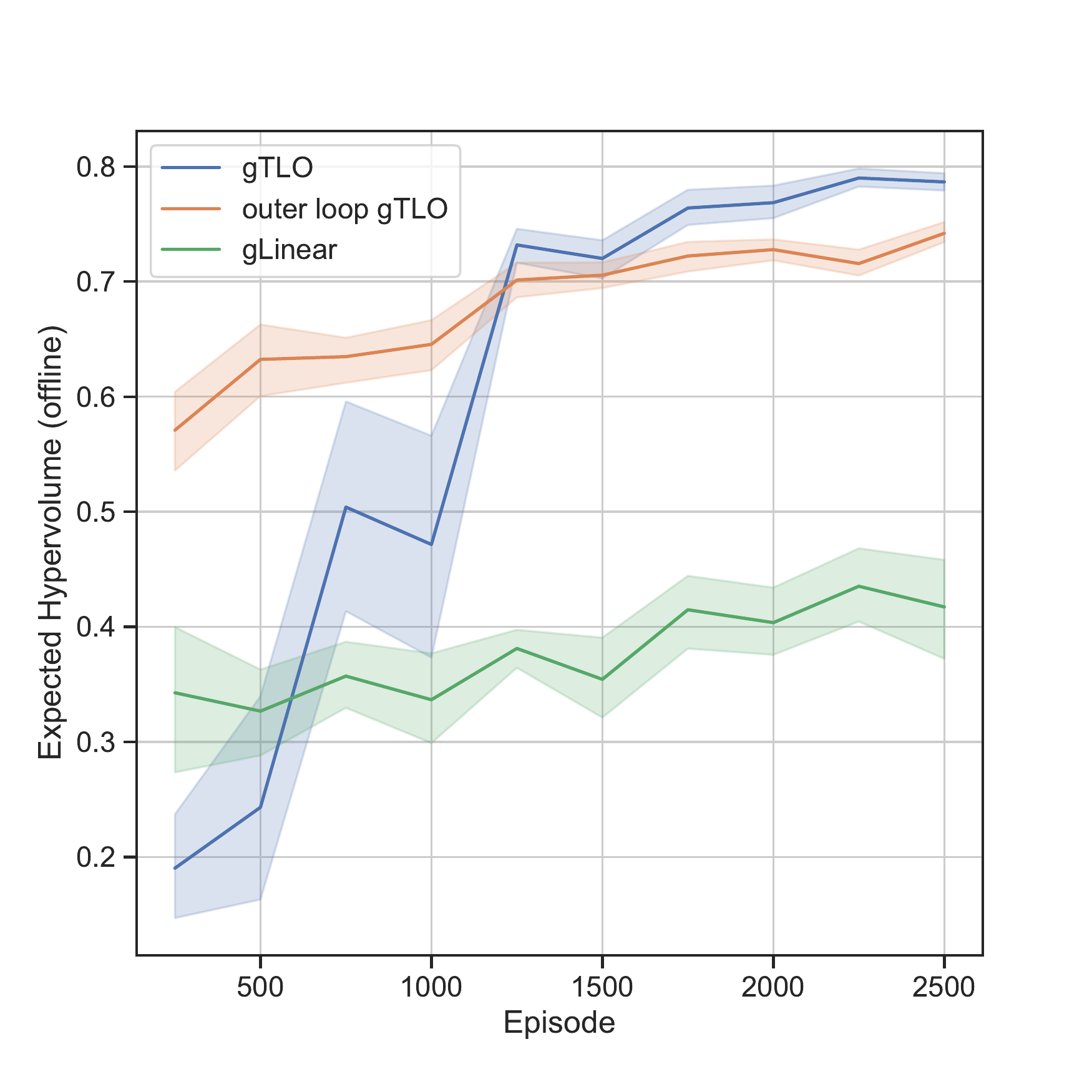}
		\caption{Expected hypervolumes over episodes ($r_{HV}=(-0.1,-0.1)$) for ten independent runs per method, calculated on evaluation phase results every 500 episodes. The mean value and the $0.95$ confidence interval are visualized per method and evaluation phase.}
		\label{fig:dd_quant_result}
	\end{figure}
	
	\begin{figure*}
		\centering
		\subfloat[gTLO solutions (after $2,500$ episodes)]{\includegraphics[width=\textwidth]{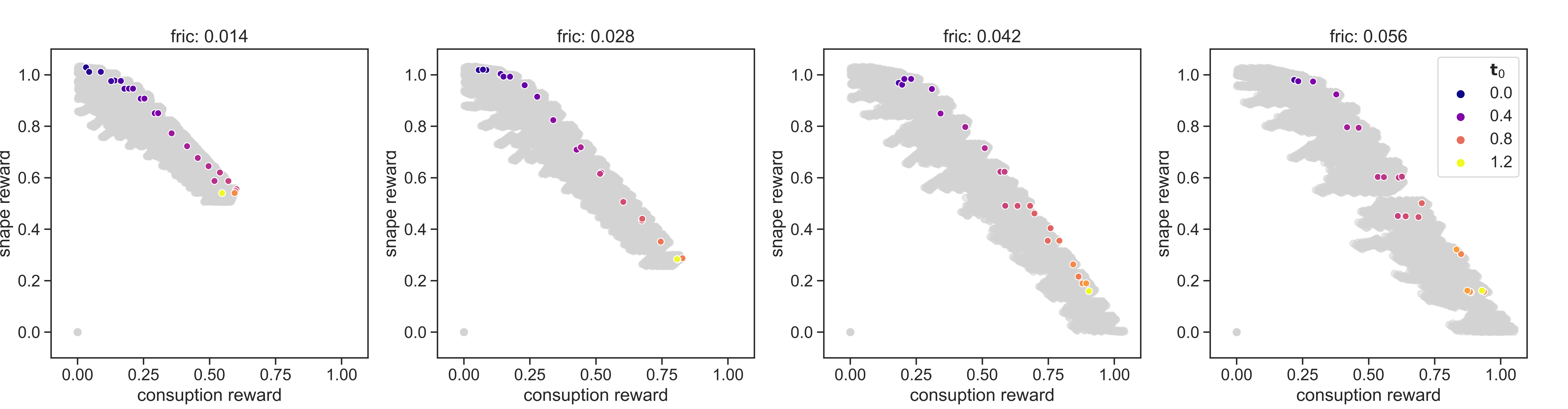}}\\
		\subfloat[\textit{outer loop gTLO} solutions (after $2,500$ episodes)]{\includegraphics[width=\textwidth]{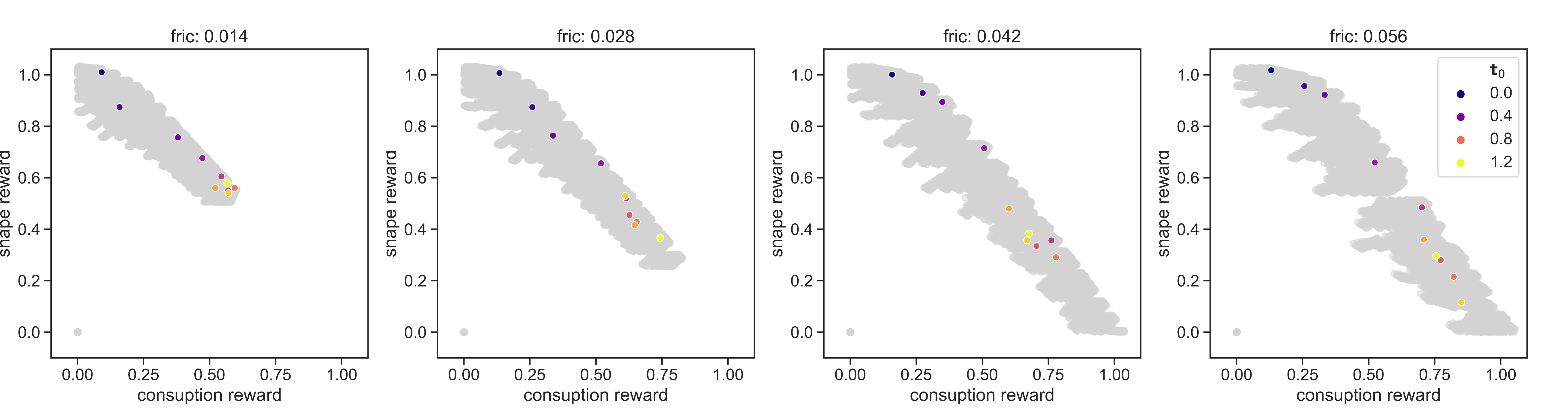}}\\
		\subfloat[gLinear solutions (after $10,000$ episodes)]{\includegraphics[width=\textwidth]{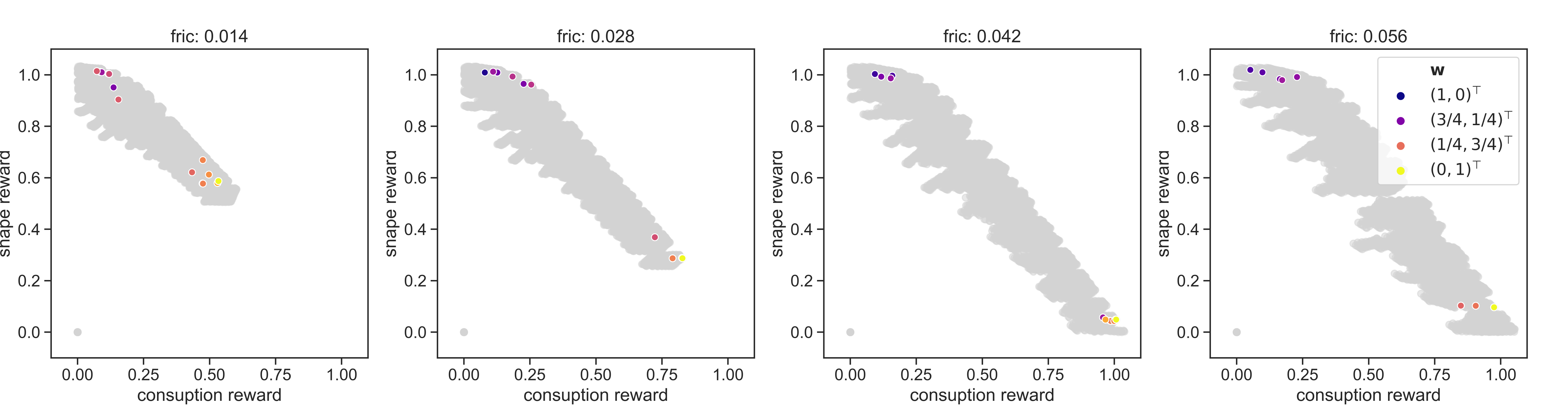}}
		\caption{Solutions found by \textit{gTLO}, \textit{outer loop gTLO} and \textit{gLinear} for various friction coefficients $\mu$ and preference weights (color coded) after $2,500$, $2,500$ and $10,000$ episodes respectively. For each friction, the attainable returns are visualized as per solution grey dots.}
			\label{fig:dd_qual_result}
	\end{figure*}
	
	In Fig. \ref{fig:dd_quant_result} the mean and the $0.95$ confidence interval of the expected hypervolume per evaluation phase is plotted. In early episodes the outer loop variant of \textit{gTLO} is superior. While \textit{gTLO} learns a general state representation, the outer loop variant focuses on one preference per independent model. However, \textit{gTLO} achieves the best quantitative results after approximately $1,250$ episodes. \textit{gLinear} is clearly dominated by \textit{gTLO} and its outer loop variant.
	
	Results that allow a more detailed comparison of the three methods are visualized in Fig. \ref{fig:dd_qual_result}. In each of the plots, for a combination of method and friction coefficient, attainable solutions are represented by grey dots, and the solution set attained by the respective method is represented by colored dots. The dot-color hereby encodes the preference value $\vec{t}_0\in[0, 1.2]$ for the \textit{gTLO variants} and $\vec{w}$ as defined in Section \ref{impl_details} for \textit{gLinear}. For \textit{gTLO}, \textit{outer loop gTLO}, we visualize results from the first of the $10$ independent runs, while, due to the slow convergence of \textit{gLinear}, we visualize results of an extended run consisting of $10,000$ episodes in this case in Fig. \ref{fig:dd_qual_result}(c). Results are visualized for the four most common friction coefficients\footnote{In over $80\%$ of the episodes one of these is the present friction coefficient (see Section \ref{deepdrawing_env} for details about $\mu$)} $\mu\in\{0.014, 0.028, 0.042, 0.056\}$.
	
	Fig. \ref{fig:dd_qual_result} reveals, that solutions found by \textit{gTLO} and \textit{outer loop gTLO} cover the reward range very well. As expected, solutions of the linear method are located in the convex regions near the two ends of the Pareto front, while the middle region of the Pareto front is uncovered. This again shows a major limitation of linear methods in multi-objective reinforcement learning. The \textit{gTLO} results are in general closer to the per-friction Pareto front. The plot reveals further, that the solutions found by \textit{gTLO} are much better ordered regarding the preference values than the solutions of the outer loop variant (the hypervolume metric does not take this into account). Different from the deep-sea treasure environment, here the Pareto front can not be fully achieved by the agent due to partial observability.\footnote{The agent has no information about the friction coefficient while making the first decision of each episode.}

	\section{Summary and Outlook}\label{conclusion}
	
	In this paper, we introduced generalized Thresholded Lexicographic Ordering (\textit{gTLO}), a multi-objective deep reinforcement learning algorithm that combines the advantages of generalized MORL and non-linear MORL. Generalized TLO learns multiple preference-dependent policies based on a single deep neural network, thereby transfering process knowledge between different objective preferences and scales to complex MORL problems with a huge set of optimal solutions. Due to the non-linear action selection, gTLO is not limited to solutions from the \textit{convex coverage set}.
	
	To the best of our knowledge, \textit{gTLO} is the first published generalized non-linear MORL method and is the first method that was shown to solve the image version of the original, non-convex DST benchmark to a very large extent. In the DST study results we highlight that \textit{gTLO} identifies all solutions from the Pareto front with rare exceptions, and thereby clearly outperforms non-generalizing outer loop methods that relies on the knowledge of positions of the sought solutions in the space of episode returns. 
	
	In the studies on our introduced multi-objective deep drawing optimal control environment, we show that \textit{gTLO} is capable to learn a diverse set of near-optimal solutions also in problems with many Pareto optimal solutions. In direct comparison to the outer loop variant, \textit{gTLO} provides a set of solutions near the Pareto front that are well-ordered with respect to the preference parameters. For both environments we show empirically that linear MORL methods, as expected, are not capable to identify solutions in non-convex regions of the Pareto front.
	
	Like other TLO-based methods, \textit{gTLO} is restricted to finite-horizon problems, in which the thresholded reward signals have to be zero during the episode except for the last step \cite{issabekov2012empirical}. To make the methods applicable to problems with non-zero thresholded rewards during the episode, we think it is promising to resume the early work of Geibel et al. \cite{geibel2006reinforcement} in the context of approximate TLO-based methods such as \textit{gTLO}.
	
	For evaluation, we implemented \textit{gTLO} based on vanilla \textit{Deep Q-Networks} (DQN). As it is orthogonal to those DQN extensions, \textit{gTLO} can be combined with \textit{Double Q-Learning} \cite{van2016deep}, \textit{Dueling Q-Learning} \cite{wang2016dueling} for more stable learning and \textit{Hindsight Experience Replay} \cite{schaul2015prioritized} or generalized MORL specific variants thereof such as \textit{Diverse Experience Replay} \cite{abels2019dynamic} for sample efficiency.

	Other promising opportunities for further development of \textit{gTLO} include (i) the evaluation on dynamic weight settings with correlated consecutive preference weights \cite{abels2019dynamic}, (ii) the evaluation on problems with more than two reward terms, and (iii) the extension to partially observed multi-objective Markov decision processes \cite{roijers2015point, wray2015multi}.

	\newpage
	\subsection{Code availability}\label{code}
	To simplify reproducibility of our results and subsequent research, we publish the source code of the following modules\footnote{https://github.com/johannes-dornheim/gTLO}:
	\begin{itemize}
		\item The implementation of \textit{gTLO} for two-objective environments,
		\item the deep-sea treasure environment, implemented based on the FruitAPI framework,
		\item the multi-objective deep drawing environment (requires Abaqus).
	\end{itemize}

	\clearpage
	
	{\appendix
		\section*{Equivalence of TLO formulations}\label{appendix}
		\label{app_equivalence}
		
		We state that the Thresholded Lexicographic Ordering action selection from Section \ref{TLO} is equivalent to the policy\footnote{without loss of generality, we ignore the case $Q(s,a_\text{a})=Q(s,a_\text{b})$, for actions $a_\text{a},a_\text{b}\in A$, which usually does not occur when function approximation is used.}
		\begin{equation}\label{TLO_policy_}
		\bar{\pi}_\text{TLO}\leftarrow 
		\begin{cases}
			\text{arg max}_{a\in A} \vec{Q}(s,a,\vec{t})_0, & \text{if }\abs{\hat{A}_{(\vec{t},0,s)}} = 0,\\
			\text{arg max}_{a\in \hat{A}_{(\vec{t},I,s)}} \vec{Q}(s,a,\vec{t})_I, & \text{if } \abs{\hat{A}_{(\vec{t},I,s)}} > 0,\\
			\text{arg max}_{a\in \hat{A}_{(\vec{t},i,s)}} \vec{Q}(s,a,\vec{t})_ {i+1}, & \text{otherwise,}
		\end{cases}
		\end{equation}
	
		where in the third case $i:=\max_{i}: \abs{\hat{A}_{(\vec{t},i,s)}} > 0$. The set $\hat{A}_{(\vec{t},i,s)}$, as defined in (\ref{ai_set}), stores all actions that are sufficient regarding the expected rewards and thresholds $\vec{t}_i$ up to objective $i$.
		
		From the definition of $\hat{A}$ follows that $\hat{A}_{(\vec{t},i,s)}\subseteq \hat{A}_{(\vec{t},i-1,s)}$ and $\abs{\hat{A}_{(\vec{t},i,s)}}\leq \abs{\hat{A}_{(\vec{t},i-1,s)}}$. Thus, for a given state $s$ and threshold vector $\vec{t}$ either all sets $\hat{A}_{(\vec{t},i,s)}$ for $i  \in [0,I]$  
		\begin{enumerate}[label=(\roman*)]
			\item are of cardinality $0$ (captured by the first case of (\ref{TLO_policy_})), 
			\item contain at least one element (captured by the second case of (\ref{TLO_policy_})), or
			\item there exists exactly one $i$ for which $\abs{\hat{A}_{(\vec{t},i,s)}} > 0 \land \abs{\hat{A}_{(\vec{t},i+1,s)}} = 0$ (captured by the last case in (\ref{TLO_policy_})).	
		\end{enumerate}
	
		From the definitions of $\vec{Qt}$ (\ref{qt}) and $\hat{A}_{(\vec{t},i,s)}$ (\ref{ai_set})) follows
		
		\begin{equation}
			\label{q_eq_t}
			\forall a\in \hat{A}_{(\vec{t},i,s)} : \vec{Qt}(s,a)_i = \vec{t}_i,
		\end{equation}
		
		and
		
		\begin{equation}
			\label{q_eq_q}
			\forall a\in A\setminus\hat{A}_{(\vec{t},i,s)} : \vec{Qt}(s,a)_i = \vec{Q}(s,a)_i.
		\end{equation}
		
		When substituting (\ref{q_eq_q}) into (\ref{tlo_action_selection}) for $\abs{\hat{A}_{(\vec{t},0,s)}} = 0$, it reduces to
		
		\begin{equation}
			\label{TLQ_general_case1}
			\text{sup}(a_\text{a}, a_\text{b}, s, 0) := \vec{Q}(s,a_\text{b})_0 > \vec{Q}(s,a_\text{a})_0 \lor \text{False}
		\end{equation}
		
		for $a_\text{a},a_\text{b} \in A$. This is equivalent to the order relation $\vec{Q}(s,a_\text{b})_0 > \vec{Q}(s,a_\text{a})_0$, that we use to determine $\pi_{\text{TLO}}(s)$ in the first case of (\ref{TLO_policy_}).
		%\textbf{nur für $\text{Q}(s,a',0) \neq \text{Q}(s,a,0)$... oben einflechten dass wir davon ausgehen ($\text{Q}(s,a',0) = \text{Q}(s,a,0)$ in der Praxis nicht relevant / kommt quasi nicht vor)}.
		
		Next, we assume that there exists an i with $\abs{\hat{A}_{(\vec{t},i,s)}} > 0$. By substituting (\ref{q_eq_t}) and (\ref{q_eq_q}) into (\ref{tlo_action_selection}) For any $a_\text{a}\in\hat{A}_{(\vec{t},i+1,s)},a_\text{b}\in \hat{A}_{(\vec{t},i,s)}\setminus\hat{A}_{(\vec{t},i+1,s)}$, we get
		
		\begin{equation}
			\label{TLQ_general_case2}
			\begin{split}
				\text{sup}&(a_\text{a}, a_\text{b}, s, i) := \vec{t}_i > \vec{t}_i \lor\\
				&\big[\vec{t}_i = \vec{t}_i \land (i=I \lor \text{sup}(a_\text{a}, a_\text{b}, s, i+1))\big],
			\end{split}
		\end{equation}
		
		where
		
		\begin{equation}
			\label{TLQ_general_case2_}
			\begin{split}
				\text{sup}(a_\text{a}, a_\text{b}, s, i+1) := \vec{t}_{i+1} > \vec{Q}(s,a_\text{b})_{i+1} \lor [...]
			\end{split}
		\end{equation}
		
		is true. It can be seen that $\text{sup}(a_\text{a}, a_\text{b}, s, i)$ is fulfilled for any $a_\text{a},a_\text{b}$ with $a_\text{a}\in\hat{A}_{(\vec{t},i+1,s)}$ and $a_\text{b}\in \hat{A}_{(\vec{t},i,s)}\setminus\hat{A}_{(\vec{t},i+1,s)}$. The action $\pi_{\text{TLO}}$ subsequentely is either an element of $\hat{A}_{(\vec{t},I,s)}$ if $\abs{\hat{A}_{(\vec{t},I,s)}}>0$ (the second case in (\ref{TLO_policy_}))\footnote{Mind that $\hat{A}_{(\vec{t},I,s)}=\hat{A}_{(\vec{t},I-1,s)}$, because by definition $t_I=:\infty$} or an element of $\hat{A}_{(\vec{t},i,s)}$, where $\abs{\hat{A}_{(\vec{t},i,s)}} > 0$ and $\abs{\hat{A}_{(\vec{t},i+1,s)}} = 0$ (the third case of (\ref{TLO_policy_})).
		
		For $k=I$ in the case $\abs{\hat{A}_{(\vec{t},I,s)}}>0$ and $k=i+1$ in the case $\abs{\hat{A}_{(\vec{t},i,s)}} > 0\land \abs{\hat{A}_{(\vec{t},i+1,s)}} = 0$ we now can see, that applying $\text{sup}(a_\text{a}, a_\text{b}, s, k)$ is equivalent to an ordering of the values in $\hat{A}_{(\vec{t},I,s)}$ and in $\hat{A}_{(\vec{t},i,s)}$ respectively:
		
		\begin{equation}
			\label{TLQ_general_case23}
			\text{sup}(a_\text{a}, a_\text{b}, s, k) := \vec{Q}(s,a_\text{b})_k > \vec{Q}(s,a_\text{a})_k \lor \text{False}.
		\end{equation}

	}

	%\bibliography{IEEEabrv, ieee_paper}
	{
		\bibliographystyle{IEEEtran}
		\bibliography{ieee_paper}
	}

	\newpage
	
	% your own custom macro containing the $\backslash${\tt{includegraphics}} command to make things
	%\bf{If you include a photo:}\vspace{-33pt}
	%\begin{IEEEbiography}[{\includegraphics[width=1in,height=1.25in,clip,keepaspectratio]{fig1}}]{Michael Shell}
	%Use $\backslash${\tt{begin\{IEEEbiography\}}} and then for the 1st argument use $\backslash${\tt{includegraphics}} to declare and link the author photo.
	%Use the author name as the 3rd argument followed by the biography text.
	%\end{IEEEbiography}

	\vfill
	
\end{document}